\documentclass[lettersize,journal]{IEEEtran}
\usepackage{amsmath,amsfonts}
\usepackage{algorithmic}
\usepackage{algorithm}
\usepackage{array}
\usepackage{bm}
\usepackage{stfloats}
\usepackage{multirow}
\usepackage{booktabs}
\usepackage{textcomp}
\usepackage{stfloats}
\usepackage{url}
\usepackage{verbatim}
\usepackage{graphicx}
\usepackage{subcaption}
\usepackage{cite}
\usepackage{subfloat}
\newcommand{\wh}[1]{\textcolor{black}{#1}}
\newcommand{\nzk}[1]{\textcolor{black}{#1}}
\newcommand{\revise}[1]{\textcolor{black}{#1}}

\usepackage{color} 
\usepackage{hyperref}
\usepackage{amssymb}
\usepackage{rotating}

\usepackage[detect-all]{siunitx}
\usepackage{etoolbox}

\begin{document}

\title{Opinion-Unaware Blind Image Quality Assessment using Multi-Scale Deep Feature Statistics}

\author{
    Zhangkai Ni,~\IEEEmembership{Member,~IEEE}, 
    Yue Liu,
    Keyan Ding,
    Wenhan Yang,~\IEEEmembership{Member,~IEEE}, \\
    Hanli Wang,~\IEEEmembership{Senior Member,~IEEE}, 
    Shiqi Wang,~\IEEEmembership{Senior Member,~IEEE} 

\thanks{This work was supported in part by the National Natural Science Foundation of China under Grant 62201387, Grant 62371343, and Grant 62301480, in part by the Shanghai Pujiang Program under Grant 22PJ1413300, and in part by the Fundamental Research Funds for the Central Universities. 
\emph{(Corresponding authors: Hanli Wang and Keyan Ding)}
}
\thanks{Zhangkai Ni and Hanli Wang are with the Department of Computer Science and Technology, Key Laboratory of Embedded System and Service Computing (Ministry of Education), and Shanghai Institute of Intelligent Science and Technology, Tongji University, Shanghai 200092, China (e-mail: zkni@tongji.edu.cn; hanliwang@tongji.edu.cn).}
\thanks{Yue Liu and Shiqi Wang are with the Department of Computer Science, City University of Hong Kong, Hong Kong 999077 (e-mail: yliu724-c@my.cityu.edu.hk; shiqwang@cityu.edu.hk).}
\thanks{Keyan Ding is with ZJU-Hangzhou Global Scientific and Technological Innovation Center, Zhejiang University, Hangzhou, Zhejiang 311200, China (e-mail: dingkeyan@zju.edu.cn).}
\thanks{Wenhan Yang is with PengCheng Laboratory, Shenzhen, Guangdong 518066, China. (e-mail: yangwh@pcl.ac.cn).}
}

\markboth{Journal of \LaTeX\ Class Files,~Vol.~14, No.~8, August~2021}%
{Shell \MakeLowercase{\textit{et al.}}: A Sample Article Using IEEEtran.cls for IEEE Journals}


\maketitle

\begin{abstract}
Deep learning-based methods have significantly influenced the blind image quality assessment (BIQA) field, however, these methods often require training using large amounts of human rating data.
In contrast, traditional knowledge-based methods are cost-effective for training but face challenges in effectively extracting features aligned with human visual perception.
\revise{To bridge these gaps, we propose integrating deep features from pre-trained visual models with a statistical analysis model into a Multi-scale Deep Feature Statistics (MDFS) model for achieving opinion-unaware BIQA (OU-BIQA), thereby eliminating the reliance on human rating data and significantly improving training efficiency.}
Specifically, we extract patch-wise multi-scale features from pre-trained vision models, which are subsequently fitted into a multivariate Gaussian (MVG) model. 
The final quality score is determined by quantifying the distance between the MVG model derived from the test image and the benchmark MVG model derived from the high-quality image set.
A comprehensive series of experiments conducted on various datasets show that our proposed model exhibits superior consistency with human visual perception compared to state-of-the-art BIQA models. Furthermore, it shows improved generalizability across diverse target-specific BIQA tasks.
\revise{Our code is available at: \url{https://github.com/eezkni/MDFS}}
\end{abstract}


\section{Introduction}

\IEEEPARstart{I}{mage} quality assessment (IQA) is a \wh{critical} and fundamental research topic in the field of computer vision due to its extensive applicability in various tasks, including image compression~\cite{li2021quality}, image super-resolution~\cite{saharia2022image}, and image enhancement~\cite{ni2020towards, wang2020exploiting}. Since the primary recipients of images are humans, the image quality scores derived from subjective quality assessment experiments are highly reliable, however, conducting such experiments is expensive and time-consuming. Therefore, numerous IQA models have been proposed in the past half-century to predict image quality that is highly consistent with the human visual system (HVS).

Existing IQA models can be divided into three categories according to the use of reference images: full-reference IQA (FR IQA), reduced-reference IQA (RR IQA), and blind IQA (BIQA).
Unlike reference-based methods, BIQA alleviates the need for direct comparisons against pristine reference images, \wh{which is} often not feasible in real-world scenarios.
Human observers \wh{are capable of assessing} image quality even in the absence of reference images, \wh{implying that} the HVS is \wh{good} at perceiving \wh{perceptual} characteristics closely associated with natural image quality.
\wh{Our research scope} is dedicated to BIQA models, with the primary objective of developing an effective and reliable method for predicting the quality score of distorted images in the absence of corresponding reference images.

BIQA methods can be divided into opinion-aware BIQA (OA-BIQA) and opinion-unaware BIQA (OU-BIQA), depending on \wh{whether relying} on subjective scores for training. 
\nzk{
Currently, most research efforts are focused on OA-BIQA models, which require training with datasets containing human-rated quality labels. However, applying deep learning to IQA is challenging due to the typically small sizes of IQA datasets~\cite{li2016blind,fang2020blind}, which increases the risk of overfitting in deep learning-based IQA models~\cite{ma2017end, guan2017visual}.
}
OU-BIQA models evaluate image quality solely based on the visual features and characteristics~\cite{wang2023visual, yang2020blind}. 
The advantage of these methods is that no subjective scoring is required during the training process, thereby reducing potential \wh{gaps} in the evaluation \wh{process} \wh{caused by} the subjectivity of different datasets, \wh{which leads to improved generalization and robustness.}
Consequently, \wh{our work emphasizes developing a robust OU-BIQA model.}
Traditional OU-BIQA methods~\cite{mittal2012making, venkatanath2015blind} have the advantage of low training costs but \wh{leave room for further improvement in performance.}
This is primarily because these approaches may not fully capture the intrinsic image characteristics aligned with human perception when assessing image quality.
\wh{Therefore, investigating the extraction of image features aligned with HVS perception is critical for the OU-BIQA algorithm.}
Deep networks enable the model to learn intrinsic representations and automatically capture important subtle image characteristics from the data~\cite{ma2019blind, wang2023deep, sim2021blind}. 
A straightforward alternative approach is to utilize a pre-trained deep network as a feature extractor, initially trained on a large-scale dataset not specifically related to the target data (\textit{i.e.}, IQA images), thus enabling the network to acquire abstract and more universally applicable features.

\wh{Considering the complementary advantages of these two kinds of methods, we aim to integrate the robustness of the traditional statistical analysis model and the universally applicable features provided by a pre-trained deep model}.
\wh{Thus, our method aims to leverage the power of DNNs to extract multi-scale features from images, and then employ statistical analysis algorithms to further analyze and aggregate the extracted features, which are integrated into the proposed Multi-scale Deep Feature Statistics (MDFS) model for OU-BIQA.}
\revise{Specifically, we first utilize a pre-trained network trained on a large-scale dataset agnostic to IQA images to generate multi-scale feature maps, and subsequently downsample and concatenate these feature maps to build feature maps with rich context and semantics.}
In the statistical data analysis stage, we calculate the mean and variance of these feature maps and utilize a multivariate Gaussian (MVG) model to model its distribution. The final MDFS index is computed by quantifying the similarity between the MVG model derived from the features of the testing image and \revise{the benchmark MVG model derived from features of a training image dataset containing only high-quality images}. 
The main contributions of our works are summarized as follows:
\begin{itemize}
\item 
We proposed the Multi-scale Deep Feature Statistic (MDFS) model for OU-BIQA, which integrates deep features extracted from DNNs into a statistical data analysis model and derives their feature distribution using an MVG model. To our knowledge, this is the first attempt to combine deep features with data distribution of traditional methods in the context of the OU-BIQA model.
\item 
We designed a statistical data analysis model to extract discriminative information from deep feature maps and integrate this information into the MVG model. 
Our work serves as a bridge to seamlessly incorporate deep features into the established traditional statistical modeling, thus paving a new way for the systematic fusion of the methods of these two categories.
\item 
The proposed MDFS outperforms state-of-the-art OU-BIQA methods in terms of cost-effectiveness during training and superior performance across diverse datasets. Moreover, it can be easily generalized to various target-specific BIQA tasks.
\end{itemize}

The rest of this paper is organized as follows. Section~\ref{sec:related} outlines the related work. Section~\ref{sec:MDFS} introduces the proposed MDFS model in detail. Section~\ref{sec:experiment} presents a comprehensive comparison and analysis of experimental results. Finally, Section~\ref{sec:conclusion} draws the conclusion.

\section{Related works}
\label{sec:related}
\revise{This section provides an overview of related BIQA methods, highlighting their scalability and flexibility advantages over various FR IQA methods specialized for different image types, such as stereoscopic images~\cite{jiang2020full, jiang2021stereoars}, retargeted images~\cite{peng2021lggd+}, and multi-exposure fused images~\cite{xu2022quality}. 
}

\subsection{Traditional BIQA Methods}
\label{sec:tro_BIQA}
Early BIQA methods are designed to assess image degradation caused by specific distortion types.
However, given the typically unknown and diverse distortion types present in a single image, there is a demand for generalized BIQA approaches capable of adapting to various complex scenarios. 
Natural scene statistics (NSS) based BIQA methods leverage the assumption that NSS is closely related to image degradation~\cite{simoncelli2001natural}, extracting diverse NSS features in spatial or transformed domains to estimate distribution disparity between reference and distorted images.

In the spatial domain, Mittal~\textit{et al.}~\cite{mittal2012making} developed the natural image quality evaluator (NIQE), where a set of image patches are selected and the corresponding statistics are regarded as quality-aware features. 
The multivariate Gaussian (MVG) model is used to estimate the global distribution of natural images, which is compared with the corresponding MVG model of the distorted image to predict the final quality score. 
Zhang~\textit{et al.}~\cite{zhang2015feature} enhanced the NIQE by incorporating color, gradient, and frequency characteristics in the feature extraction stage, resulting in the Integrated Local NIQE (ILNIQE).
To reduce the feature dimension in ILNIQE, Liu~\textit{et al.}~\cite{liu2019unsupervised} implies sparse representation in the proposed structure, naturalness, and perception quality-driven NIQE (SNP-NIQE). 
They further proposed the natural scene statistics and perceptual characteristics-based quality index (NPQI)~\cite{liu2020blind} by applying the local binary pattern map and the locally mean subtracted and contrast normalized (MSCN) coefficients of the image to extract the NSS features. 
Furthermore, Xue \textit{et al.}~\cite{xue2013learning} proposed a quality-aware clustering (QAC) method by learning a set of quality centroids, which is used to estimate the quality of each patch of the input image. 
Later, Venkatanath~\textit{et al.}~\cite{venkatanath2015blind} introduced a visual attention strategy and developed a perception-based image quality evaluator (PIQE) to improve the assessment accuracy. 
Wu~\textit{et al.}~\cite{wu2015highly} designed a local pattern statistics index (LPSI) by modifying the statistics of the feature extracted from the local binary pattern.

In addition to directly extracting NSS features in the spatial domain, BIQA methods have explored frequency domain features, such as those in the discrete cosine transform (DCT) domain and wavelet domain.
Moorthy~\textit{et al.}~\cite{moorthy2011blind} modeled the wavelet coefficients and further identified the distortion type of the image using a support vector machine (SVM) to produce the final quality score. 
Saad~\textit{et al.}~\cite{saad2012blind} applied a generalized Gaussian distribution (GGD) model to predict image quality based on features extracted from the DCT coefficients. 
Furthermore, Wu~\textit{et al.}~\cite{wu2015blind} combined features from multiple domains and color channels in the proposed type classification and label transfer (TCLT) model to estimate perceptual image quality.

\begin{figure*}[t]
    \begin{center}
     \includegraphics[width=0.90\linewidth]{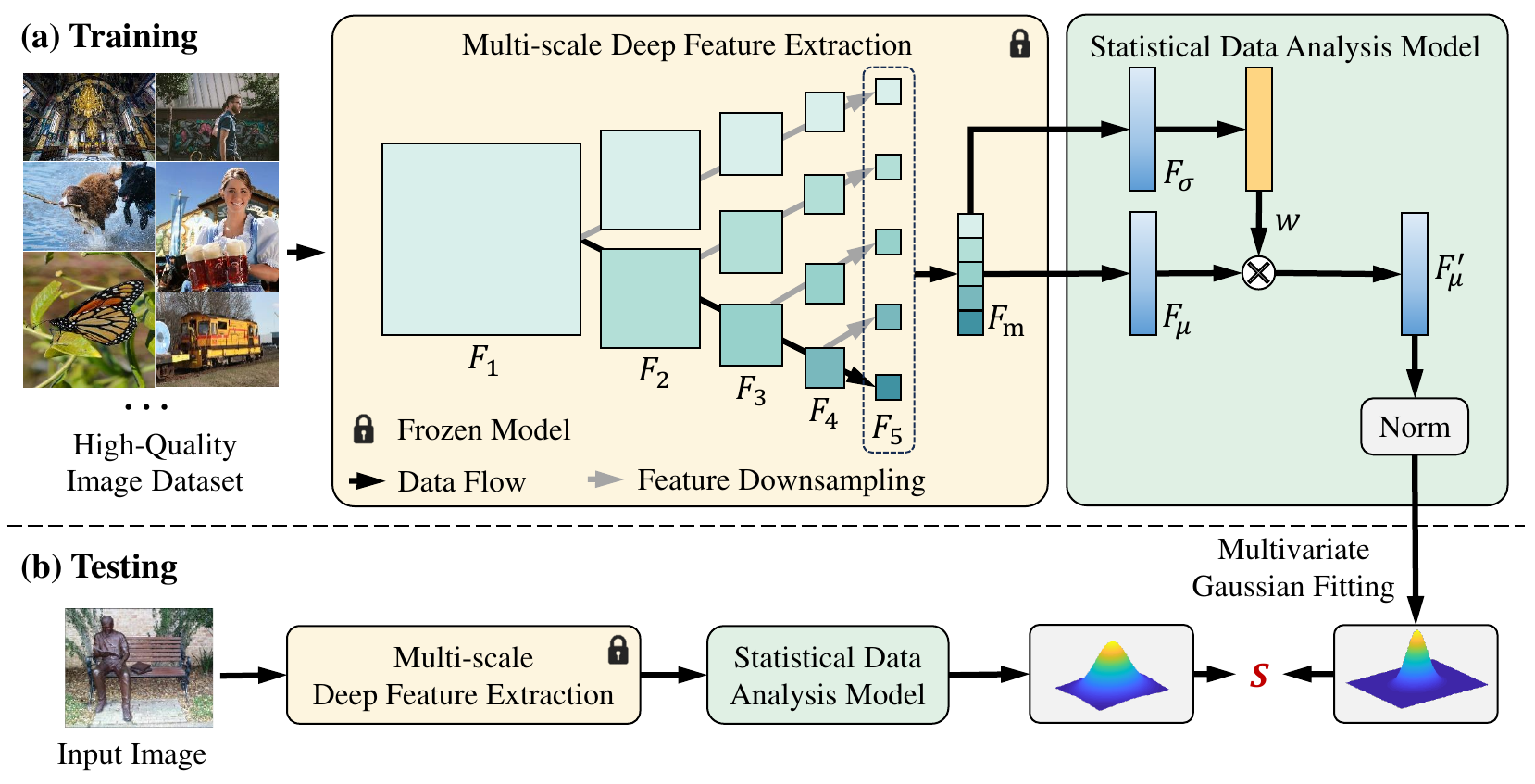}
    \end{center}
     \caption{
     \revise{     Overview of the proposed MDFS model:
    (a) Training phase: This process involves fitting a benchmark multivariate Gaussian (MVG) model from a set of high-quality images, including a frozen multi-scale deep feature extraction module (\textit{e.g.}, ResNet, VGG, and EfficientNet), a statistical data analysis model, and an MVG fitting model.
    (b) Testing phase: The process of assessing the quality of a test image involves calculating the final quality score by measuring the distance between an MVG model fitted using the test image features and the benchmark MVG model obtained in the training phase.}
    }
\label{fig:Overview}
\end{figure*}

\subsection{Learning-based BIQA Methods}
\label{sec:lb_BIQA}
In recent years, convolutional neural networks (CNNs) have made significant strides in the field of BIQA, with techniques including graph convolutional neural networks~\cite{sun2022graphiqa}, continual learning approaches~\cite{zhang2022continual}, and attention mechanisms~\cite{su2020blindly, zhu2022blind}.
These methods can be classified into supervised (opinion-aware) and unsupervised (opinion-unaware) BIQA algorithms, based on the availability of subjective scores for training.
For opinion-aware OA-BIQA methods, Kim~\textit{et al.}~\cite{kim2016fully} applied an FR IQA method as an auxiliary task to guide the network to learn quality map, where intermediate features are utilized to predict the final quality score. Guan \textit{et al.}~\cite{guan2017visual} proposed to extract features of each patch and then use a local regression module to generate local responses and weights, which are combined for final quality prediction. 
Ma \textit{et al.}~\cite{ma2017end} built an end-to-end network where the first sub-network classifies distortion types to assist the second sub-network in severity assessment.
To combine more IQA datasets for BIQA model training, Zhang~\textit{et al.}~\cite{zhang2022continual} introduced a new head for new datasets in the training process, along with existing heads for predicting image quality, where k-means clustering is used to generate the weighting map.
Besides, graph representation has been applied in BIQA~\cite{sun2022graphiqa}, where the node and edge of the learned graph represent the distortion types and distortion levels.
Recently, Su \textit{et al.}~\cite{su2020blindly} took the attention mechanism into \wh{the} BIQA model for better content understanding. Yang \textit{et al.}~\cite{yang2022maniqa} also introduced a transformer in the BIQA task.

OA-BIQA methods can achieve state-of-the-art performance on specific datasets with the same distribution as the training dataset, but their generalizability to new datasets has been discredited due to potential overfitting issues ~\cite{zhang2022continual}.
In contrast, OU-BIQA algorithms make it easier to establish the training datasets and can adapt to adjustments based on new, unlabeled datasets, making them more useful for various applications. 
As a result, various researchers have shifted their focus toward OU-BIQA algorithms. 
Ma~\textit{et al.}~\cite{ma2019blind} proposed to generate a set of distorted image pairs, where higher quality images are identified by multiple IQA methods. This generated dataset is later used for image quality estimation. 
In their subsequent work~\cite{ma2017dipiq}, they further improved this model by combining the training data into a quality discriminable image pair (DIP) format, which is subsequently fed into a pairwise learning-to-rank algorithm for quality measurement.
Recently, Chen~\textit{et al.}~\cite{chen2022spiq} proposed a self-supervised strategy, where the quality-aware information is learned from a patch prediction framework based on contrastive learning.
Babu~\textit{et al.}~\cite{babu2023no} proposed a self-supervised method, where mutual information bounds are used to separate content information from image patches, focusing on content-independent image quality.

\section{Multi-scale Deep Feature Statistic Model}
\label{sec:MDFS}

\subsection{Overview}
\label{sec:overview}

Previous OA-BIQA methods suffer from the risk of overfitting due to the limited data while conventional OU-BIQA fails to obtain satisfactory performance caused by the lack of intrinsic image features.
Our goal is to build a robust and efficient OU-IQA method by applying statistical analysis models for universally applicable features extracted from a pre-trained deep network to simultaneously inherit high performance in terms of robustness and training efficiency.
We innovatively leverage multi-scale features learned by a deep neural network to fit a multivariate Gaussian (MVG) model.
The framework of our proposed MDFS for OU-BIQA is shown in Figure~\ref{fig:Overview}, which consists of training and testing phases, outlined as follows:
\begin{itemize}
    \item In the training phase, image features are initially extracted from a pre-trained neural network using \textbf{Multi-Scale Deep Feature Extraction}, followed by analysis with a \textbf{Statistical Data Analysis Model}, \revise{and then fitted to the benchmark \textbf{Multivariate Gaussian} (MVG) model}.
    \item \revise{In the testing phase, the MVG model of the distorted image is obtained through a process similar to the training phase. The final quality score of the test image is determined by evaluating the distance between the MVG model of the test image and the benchmark MVG model.}
\end{itemize}  
Each component of the two phases will be detailed in the following subsections, respectively.

\subsection{Multi-scale Deep Feature Extraction}
\label{sec:feature_extraction}
In recent years, deep learning has significantly improved the accuracy and efficiency of numerous computer vision tasks, with various classic network architectures proposed, such as ViT~\cite{dosovitskiy2020image}, VGG~\cite{simonyan2014very}, ConvNet~\cite{liu2022convnet}, ResNet~\cite{he2016deep}, and EfficientNet~\cite{tan2019efficientnet}, serving as the foundation for various downstream tasks.
The multi-scale deep feature extraction modal extracts multi-scale features in a pyramid form for better feature representation. 
Specifically, given an input image $I_{in}$, the pre-trained vision modal (\textit{e.g.}, VGG, ResNet, and EfficientNet) is first used to extract image features. The outputs of the first layer to the fifth layer of the model are specially extracted and represented as $\bm{F}_i (i = 1, 2, 3, 4, 5)$.
Features of the first three layers are downsampled to unify multiple features of different scales, which is defined as,
\begin{equation}
\bm{F}_m = \bm{F}_5 \circ D(\bm{F}_4 \circ {\rm D}(\bm{F}_3 \circ {\rm D}(\bm{F}_2 \circ {\rm D}(\bm{F}_1)))),
\end{equation}
where $D(\cdot)$ indicates the downsampling operation and $\circ$ is concatenation. 
\revise{The downsampling is implemented using the same non-learnable convolutional layer with a stride of 2 to halve the scale of feature maps, and reflection padding is applied to mitigate edge artifacts. Note that the multi-scale deep feature extraction is not learnable and the kernel used in the downsampling layer remains fixed during the training and testing stage.}

\subsection{Statistical Data Analysis Model}
\label{sec:data_analysis}
In traditional BIQA methods~\cite{mittal2012making,liu2020blind}, NSS features are widely explored to \wh{identify} a perceptual feature extraction function that closely approximates the sensitivity of the HVS. However, how to perform statistical analysis for learning-based features has received little attention.
In this section, a statistical data analysis model is proposed for further data distillation and analysis based on extracted deep features.
In this model, the non-learnable convolution operation is utilized to estimate the local mean value of the feature map, which is subsequently screened before normalization. 
Specifically, given an input deep feature $\bm{F}_m\in\mathbb{R}^{C\times H\times W}$, where $C$, $H$, and $W$ denotes the number of channels, height, and width respectively, a Gaussian filter \revise{with dynamic window size $s_w$} is applied on the feature to estimate the local mean value of the feature map. 
Therefore, statistical features, including the mean and standard deviation of $\bm{F}_m$, are generated as follows,
\begin{equation}
\bm{F}_\mu = {\rm conv}(\bm{F}_m \revise{, s_w}),
\label{eq:conv}
\end{equation}
\begin{equation}
\bm{F}_\sigma ={\rm mean_c}\sqrt{{\rm conv}(\bm{F}_m^2\revise{, s_w})},
\end{equation}
where ${\rm conv(\cdot)}$ and ${\rm mean_c(\cdot)}$ denote the convolutional operation with the stride of 1 and the averaging operation in the channel dimension, respectively. 
To enable the filter to adapt to feature maps of different sizes, a dynamic window size calculation method is employed as follows,
\begin{equation}
s_w = {\rm max}\left(3, 1+2\cdot\left ({\rm min}(H, W)//2^k\right)\right),
\label{eq:win}
\end{equation}
where $//$ represents the remainder operation and $k$ is empirically set to 5 in this work.
\revise{Since the feature map is obtained by concatenating features from five different scales, we normalize $\bm{F}_{\mu}$ along the channel dimension of each layer to obtain the final features for:
\begin{equation}
\bm{F}_{\mu}^{’}=\text{norm}_{c}(\bm{F}_{\mu}),
\end{equation}
where $\text{norm}_{c}(\cdot)$ represents the normalization operation along the channel dimension.}

\revise{Considering that the HVS is more sensitive to information with higher contrast~\cite{fu2018screen}, we utilize standard deviation as a measure of contrast and highlight local regions exhibiting greater standard deviation. The weighting map is defined as:
\begin{equation}
\bm{w} = \frac{1}{1+e^{-(\bm{F}_\sigma -\bm{F}_\sigma^\mu)/(\bm{F}_\sigma^\sigma + \delta)}},
\end{equation}
where $\bm{F}_\sigma^\mu $ and $\bm{F}_\sigma^\sigma $ are the mean and standard deviation of the $\bm{F}_\sigma$, respectively. 
$\delta $ is a small positive number (\textit{i.e.}, $\delta=1 \times e^{-12}$) to avoid the denominator being equal to 0. 
The weighting map of the distorted image is adopted in the testing stage to calculate the weighted quality score.}

\begin{table*}[t]
  \centering
  \caption{Performance comparisons of different OU-BIQA models on ten public datasets. The top three performers are marked in bold~\textcolor[rgb]{ 1,  0,  0}{\textbf{red}}, \textcolor[rgb]{ 0,  0,  1}{\textbf{blue}}, and~\textbf{black}, respectively.}
    \begin{tabular}{cc|cccccccccc}
    \toprule
    \toprule
      Metrics    & \multicolumn{1}{c|}{Datasets} & NIQE  & QAC   & PIQE  & LPSI  & ILNIQE & dipIQ & SNP-NIQE & NPQI  & ContentSep & MDFS (Ours) \\
    \midrule 
    \multirow{12}[6]{*}{\begin{sideways}SROCC\end{sideways}} 
   & LIVE  & 0.9062 & 0.8683 & 0.8398 & 0.8181 & 0.8975 & \textcolor[rgb]{ 1,  0,  0}{\textbf{0.9378}} & 0.9073 & \textbf{0.9108} & 0.7478 & \textcolor[rgb]{ 0,  0,  1}{\textbf{0.9361}} \\
   & CSIQ  & 0.6191 & 0.4804 & 0.5120 & 0.5218 & \textcolor[rgb]{ 1,  0,  0}{\textbf{0.8045}} & 0.5191 & 0.6090 & \textbf{0.6341} & 0.5871 & \textcolor[rgb]{ 0,  0,  1}{\textbf{0.7774}} \\
   & TID2013 & 0.3106 & 0.3719 & 0.3636 & 0.3949 & \textcolor[rgb]{ 0,  0,  1}{\textbf{0.4938}} & \textbf{0.4377} & 0.3329 & 0.2804 & 0.2530 & \textcolor[rgb]{ 1,  0,  0}{\textbf{0.5363}} \\
   & KADID & 0.3779 & 0.2394 & 0.2372 & 0.1478 & \textcolor[rgb]{ 0,  0,  1}{\textbf{0.5406}} & 0.2977 & 0.3719 & 0.3909 & \textbf{0.5060} & \textcolor[rgb]{ 1,  0,  0}{\textbf{0.5983}} \\
   & MDLIVE & 0.7728 & 0.4116 & 0.3862 & 0.2717 & \textcolor[rgb]{ 1,  0,  0}{\textbf{0.8778}} & 0.6678 & \textbf{0.7822} & \textcolor[rgb]{ 0,  0,  1}{\textbf{0.8100}} & 0.4285 & 0.7579 \\
   & MDIVL & 0.5656 & 0.5524 & 0.5319 & 0.5736 & 0.6237 & \textcolor[rgb]{ 0,  0,  1}{\textbf{0.7131}} & \textbf{0.6252} & 0.6139 & 0.2582 & \textcolor[rgb]{ 1,  0,  0}{\textbf{0.7890}} \\
   & KonIQ & 0.5300 & 0.3397 & 0.2452 & 0.2239 & 0.5057 & 0.2375 & \textbf{0.6284} & 0.6132 & \textcolor[rgb]{ 0,  0,  1}{\textbf{0.6401}} & \textcolor[rgb]{ 1,  0,  0}{\textbf{0.7333}} \\
   & CLIVE & 0.4495 & 0.2258 & 0.2325 & 0.0832 & 0.4393 & 0.2089 & 0.4654 & \textbf{0.4752} & \textcolor[rgb]{ 1,  0,  0}{\textbf{0.5060}} & \textcolor[rgb]{ 0,  0,  1}{\textbf{0.4821}} \\
   & CID2013 & 0.6589 & 0.0299 & 0.0448 & 0.3229 & 0.3062 & 0.3776 & \textbf{0.7159} & \textcolor[rgb]{ 0,  0,  1}{\textbf{0.7698}} & 0.6116 & \textcolor[rgb]{ 1,  0,  0}{\textbf{0.8571}} \\
   & SPAQ  & 0.3105 & 0.4397 & 0.2317 & 0.0001 & \textbf{0.6959} & 0.2189 & 0.5402 & 0.5999 & \textcolor[rgb]{ 0,  0,  1}{\textbf{0.7084}} & \textcolor[rgb]{ 1,  0,  0}{\textbf{0.7408}} \\
   \cmidrule{2-12} 
   & $AVG_D$ & 0.5501 & 0.3959 & 0.3625 & 0.3358 & \textcolor[rgb]{ 0,  0,  1}{\textbf{0.6185}} & 0.4616 & 0.5978 & \textbf{0.6098} & 0.5247 & \textcolor[rgb]{ 1,  0,  0}{\textbf{0.7208}} \\
   & $AVG_W$ & 0.4226 & 0.3562 & 0.2706 & 0.1760 & \textcolor[rgb]{ 0,  0,  1}{\textbf{0.5851}} & 0.2987 & 0.5164 & 0.5322 & \textbf{0.5815} & \textcolor[rgb]{ 1,  0,  0}{\textbf{0.6854}} \\
    \midrule
    \multirow{12}[6]{*}{\begin{sideways}KROCC\end{sideways}} 
    & LIVE  & 0.7275 & 0.6738 & 0.6367 & 0.6175 & 0.7123 & \textcolor[rgb]{ 1,  0,  0}{\textbf{0.7806}} & 0.7353 & \textbf{0.7421} & 0.5456 & \textcolor[rgb]{ 0,  0,  1}{\textbf{0.7709}} \\
   &  CSIQ  & 0.4520 & 0.3452 & 0.3687 & 0.3736 & \textcolor[rgb]{ 1,  0,  0}{\textbf{0.6109}} & 0.3963 & 0.4492 & \textbf{0.4819} & 0.4057 & \textcolor[rgb]{ 0,  0,  1}{\textbf{0.5823}} \\
   &  TID2013 & 0.2114 & 0.2575 & 0.2554 & 0.2734 & \textcolor[rgb]{ 0,  0,  1}{\textbf{0.3491}} & \textbf{0.3016} & 0.2285 & 0.1960 & 0.1676 & \textcolor[rgb]{ 1,  0,  0}{\textbf{0.3824}} \\
   &  KADID & 0.2624 & 0.1660 & 0.1657 & 0.1004 & \textcolor[rgb]{ 0,  0,  1}{\textbf{0.3808}} & 0.2133 & 0.2584 & 0.2732 & \textbf{0.3426} & \textcolor[rgb]{ 1,  0,  0}{\textbf{0.4238}} \\
   &  MDLIVE & 0.5799 & 0.2903 & 0.2715 & 0.2071 & \textcolor[rgb]{ 1,  0,  0}{\textbf{0.6880}} & 0.4872 & \textbf{0.5923} & \textcolor[rgb]{ 0,  0,  1}{\textbf{0.6196}} & 0.2917 & 0.5623 \\
   &  MDIVL & 0.3934 & 0.3751 & 0.3642 & 0.3998 & 0.4383 & \textcolor[rgb]{ 0,  0,  1}{\textbf{0.5034}} & \textbf{0.4444} & 0.4359 & 0.1641 & \textcolor[rgb]{ 1,  0,  0}{\textbf{0.5911}} \\
   &  KonIQ & 0.3679 & 0.2302 & 0.1649 & 0.1504 & 0.3504 & 0.1594 & \textbf{0.4434} & 0.4310 & \textcolor[rgb]{ 0,  0,  1}{\textbf{0.4529}} & \textcolor[rgb]{ 1,  0,  0}{\textbf{0.5344}} \\
   &  CLIVE & 0.3064 & 0.1514 & 0.1561 & 0.0523 & 0.2984 & 0.1395 & 0.3162 & \textbf{0.3256} & \textcolor[rgb]{ 1,  0,  0}{\textbf{0.3450}} & \textcolor[rgb]{ 0,  0,  1}{\textbf{0.3274}} \\
   &  CID2013 & 0.4675 & 0.0178 & 0.0394 & 0.2168 & 0.2100 & 0.2614 & \textbf{0.5156} & \textcolor[rgb]{ 0,  0,  1}{\textbf{0.5655}} & 0.4378 & \textcolor[rgb]{ 1,  0,  0}{\textbf{0.6706}} \\
   &  SPAQ  & 0.2059 & 0.3001 & 0.1560 & 0.0006 & \textbf{0.4930} & 0.1454 & 0.3686 & 0.4137 & \textcolor[rgb]{ 0,  0,  1}{\textbf{0.5069}} & \textcolor[rgb]{ 1,  0,  0}{\textbf{0.5347}} \\
   \cmidrule{2-12}
   &  $AVG_D$ & 0.3974 & 0.2807 & 0.2579 & 0.2392 & \textcolor[rgb]{ 0,  0,  1}{\textbf{0.4531}} & 0.3388 & 0.4352 & \textbf{0.4485} & 0.3660 & \textcolor[rgb]{ 1,  0,  0}{\textbf{0.5380}} \\
   &  $AVG_W$ & 0.2932 & 0.2456 & 0.1869 & 0.1216 & \textcolor[rgb]{ 0,  0,  1}{\textbf{0.4145}} & 0.2093 & 0.3619 & 0.3749 & \textbf{0.4075} & \textcolor[rgb]{ 1,  0,  0}{\textbf{0.4966}} \\
    \midrule
    \multirow{12}[6]{*}{\begin{sideways}PLCC\end{sideways}} 
   & LIVE  & 0.9041 & 0.8625 & 0.8197 & 0.7859 & 0.9022 & \textcolor[rgb]{ 1,  0,  0}{\textbf{0.9295}} & \textbf{0.9060} & \textcolor[rgb]{ 0,  0,  1}{\textbf{0.9161}} & 0.4639 & 0.8558 \\
    &CSIQ  & 0.6901 & 0.5934 & 0.6279 & 0.6950 & \textcolor[rgb]{ 0,  0,  1}{\textbf{0.7232}} & \textbf{0.7009} & 0.6962 & 0.6479 & 0.3632 & \textcolor[rgb]{ 1,  0,  0}{\textbf{0.7907}} \\
    &TID2013 & 0.3789 & 0.4190 & 0.4615 & 0.4594 & \textcolor[rgb]{ 0,  0,  1}{\textbf{0.5090}} & \textbf{0.4746} & 0.4055 & 0.4000 & 0.2203 & \textcolor[rgb]{ 1,  0,  0}{\textbf{0.6242}} \\
    &KADID & 0.3883 & 0.3088 & 0.2887 & 0.3348 & \textcolor[rgb]{ 0,  0,  1}{\textbf{0.5341}} & 0.3832 & \textbf{0.4212} & 0.3401 & 0.3568 & \textcolor[rgb]{ 1,  0,  0}{\textbf{0.5939}} \\
    &MDLIVE & 0.8378 & 0.4149 & 0.3778 & 0.3727 & \textcolor[rgb]{ 1,  0,  0}{\textbf{0.8923}} & 0.7241 & \textcolor[rgb]{ 0,  0,  1}{\textbf{0.8525}} & \textbf{0.8454} & 0.3524 & 0.8226 \\
    &MDIVL & 0.5650 & 0.5713 & 0.5142 & 0.5715 & 0.5697 & \textcolor[rgb]{ 0,  0,  1}{\textbf{0.7252}} & \textbf{0.6393} & 0.6013 & 0.2311 & \textcolor[rgb]{ 1,  0,  0}{\textbf{0.7953}} \\
    &KonIQ & 0.4835 & 0.2906 & 0.2061 & 0.1064 & 0.4963 & 0.3773 & \textbf{0.6222} & 0.6139 & \textcolor[rgb]{ 0,  0,  1}{\textbf{0.6274}} & \textcolor[rgb]{ 1,  0,  0}{\textbf{0.7123}} \\
    &CLIVE & 0.4939 & 0.2841 & 0.3144 & 0.2521 & 0.5033 & 0.3163 & \textcolor[rgb]{ 0,  0,  1}{\textbf{0.5199}} & 0.4920 & \textbf{0.5130} & \textcolor[rgb]{ 1,  0,  0}{\textbf{0.5364}} \\
    &CID2013 & 0.6712 & 0.0981 & 0.1072 & 0.4439 & 0.4267 & 0.3829 & \textbf{0.7260} & \textcolor[rgb]{ 0,  0,  1}{\textbf{0.7772}} & 0.6368 & \textcolor[rgb]{ 1,  0,  0}{\textbf{0.8717}} \\
    &SPAQ  & 0.2639 & 0.4497 & 0.2488 & 0.1183 & \textbf{0.6371} & 0.2239 & 0.5469 & 0.6155 & \textcolor[rgb]{ 0,  0,  1}{\textbf{0.6648}} & \textcolor[rgb]{ 1,  0,  0}{\textbf{0.7177}} \\
    \cmidrule{2-12} 
    &$AVG_D$ & 0.5677 & 0.4292 & 0.3966 & 0.4140 & 0.6194 & 0.5238 & \textcolor[rgb]{ 0,  0,  1}{\textbf{0.6336}} & \textbf{0.6249} & 0.4430 & \textcolor[rgb]{ 1,  0,  0}{\textbf{0.7321}} \\
    &$AVG_W$ & 0.4090 & 0.3735 & 0.2914 & 0.2441 & \textcolor[rgb]{ 0,  0,  1}{\textbf{0.5661}} & 0.3697 & \textbf{0.5399} & 0.5340 & 0.5127 & \textcolor[rgb]{ 1,  0,  0}{\textbf{0.6803}} \\
    \midrule
    \multirow{12}[6]{*}{\begin{sideways}RMSE\end{sideways}} 
    &LIVE  & 11.6733 & 13.8258 & 15.6508 & 16.8932 & 11.7834 & \textcolor[rgb]{ 1,  0,  0}{\textbf{10.0761}} & \textbf{11.5643} & \textcolor[rgb]{ 0,  0,  1}{\textbf{10.9529}} & 24.2043 & 14.1344 \\
    & CSIQ  & 0.1900 & 0.2113 & 0.2043 & 0.1888 & \textcolor[rgb]{ 0,  0,  1}{\textbf{0.1813}} & \textbf{0.1873} & 0.1885 & 0.2000 & 0.2631 & \textcolor[rgb]{ 1,  0,  0}{\textbf{0.1607}} \\
    & TID2013 & 1.1472 & 1.1256 & 1.0998 & 1.1011 & \textcolor[rgb]{ 0,  0,  1}{\textbf{1.0670}} & \textbf{1.0911} & 1.1332 & 1.1362 & 1.2092 & \textcolor[rgb]{ 1,  0,  0}{\textbf{0.9685}} \\
    & KADID & 0.9977 & 1.0297 & 1.0365 & 1.0201 & \textcolor[rgb]{ 0,  0,  1}{\textbf{0.9153}} & 1.0000 & \textbf{0.9819} & 1.0181 & 1.0114 & \textcolor[rgb]{ 1,  0,  0}{\textbf{0.8710}} \\
    & MDLIVE & 10.3244 & 17.2073 & 17.5107 & 17.5497 & \textcolor[rgb]{ 1,  0,  0}{\textbf{8.5379}} & 13.0432 & \textcolor[rgb]{ 0,  0,  1}{\textbf{9.8857}} & \textbf{10.1012} & 17.6986 & 10.7534 \\
    & MDIVL & 19.7054 & 19.6015 & 20.4821 & 19.5972 & 19.6279 & \textcolor[rgb]{ 0,  0,  1}{\textbf{16.4441}} & \textbf{18.3642} & 19.0829 & 23.2351 & \textcolor[rgb]{ 1,  0,  0}{\textbf{14.4779}} \\
    & KonIQ & 0.4833 & 0.5283 & 0.5403 & 0.5741 & 0.4794 & 0.5114 & \textbf{0.4323} & 0.4359 & \textcolor[rgb]{ 0,  0,  1}{\textbf{0.4300}} & \textcolor[rgb]{ 1,  0,  0}{\textbf{0.3876}} \\
    & CLIVE & 17.6477 & 19.4601 & 19.2687 & 19.6410 & 17.5379 & 19.2545 & \textcolor[rgb]{ 0,  0,  1}{\textbf{17.3379}} & 17.6704 & \textbf{17.4226} & \textcolor[rgb]{ 1,  0,  0}{\textbf{17.1298}} \\
    & CID2013 & 16.7826 & 22.5312 & 22.5098 & 20.2875 & 20.4756 & 20.9148 & \textbf{15.5695} & \textcolor[rgb]{ 0,  0,  1}{\textbf{14.2467}} & 17.4576 & \textcolor[rgb]{ 1,  0,  0}{\textbf{11.0931}} \\
    & SPAQ  & 20.1607 & 18.6684 & 20.2439 & 20.7546 & \textbf{16.1107} & 20.3706 & 17.4992 & 16.4737 & \textcolor[rgb]{ 0,  0,  1}{\textbf{15.6132}} & \textcolor[rgb]{ 1,  0,  0}{\textbf{14.5551}} \\
   \cmidrule{2-12} 
    & $AVG_D$ & 9.9112 & 11.4189 & 11.8547 & 11.7607 & 9.6716 & 10.2893 & \textbf{9.2957} & \textcolor[rgb]{ 0,  0,  1}{\textbf{9.1318}} & 11.8545 & \textcolor[rgb]{ 1,  0,  0}{\textbf{8.4532}} \\
    & $AVG_W$ & 7.7271 & 7.5637 & 8.0693 & 8.2122 & \textcolor[rgb]{ 0,  0,  1}{\textbf{6.5589}} & 7.8258 & 6.8883 & \textbf{6.6031} & 6.8263 & \textcolor[rgb]{ 1,  0,  0}{\textbf{5.9160}} \\
    \bottomrule
    \bottomrule
    \end{tabular}%
  \label{tab:overall}%
\end{table*}%

\subsection{Multivariate Gaussian Model}
\label{sec:mvg}
The MVG model has been extensively utilized to model the joint probability distribution of a vector of random variables, with each variable following a normal distribution. 
Herein, we use the MVG model to estimate the joint probability distribution of a set of training images. Let $\overrightarrow{X} = [x_1, \cdot\cdot\cdot, x_n], (n = 1,2,3, \cdot\cdot\cdot)$ represent the statistical features obtained from $n$ high-quality images.
Assuming that these features represent independent samples from an $l$-dimensional MVG distribution, the MVG model learned through Maximum Likelihood Estimation (MLE) can be expressed as follows:
\begin{equation}
p(\overrightarrow{X}) = \frac{1}{\sqrt{(2\pi)^l|\sum|}}\cdot e^{-\frac{1}{2}(\overrightarrow{X}-\overrightarrow{\mu})^{\rm T}\sum^{-1}(\overrightarrow{X}-\overrightarrow{\mu})},
\end{equation}
where $l$ is the dimension of the learned statistical features, while $\overrightarrow{\mu}$ and $\sum$ denote the mean and covariance matrix of the estimated MVG model, respectively.

\begin{table*}[t]
  \centering
  \caption{SROCC and KROCC comparisons of various IQA models under different distortion types on the TID2013 \wh{dataset}. 
  }
    \begin{tabular}{c|cccccccccc}
    \toprule
    \toprule
    \multirow{2}{*}{Distortions} &  \multicolumn{10}{c}{SROCC $\uparrow$}  \\
    \cmidrule{2-11} & NIQE & QAC & PIQE & LPSI & ILNIQE & dipIQ & SNP-NIQE & NPQI & ContentSep & {MDFS (Ours)} \\
    \midrule
    ANG   & \multicolumn{1}{c}{0.8187} & \multicolumn{1}{c}{0.7427} & \multicolumn{1}{c}{0.8555} & \multicolumn{1}{c}{0.7692} & \multicolumn{1}{c}{\textcolor[rgb]{ 0,  0,  1}{\textbf{0.8767}}} & \multicolumn{1}{c}{\textbf{0.8653}} & \multicolumn{1}{c}{\textcolor[rgb]{ 1,  0,  0}{\textbf{0.8855}}} & \multicolumn{1}{c}{0.6257} & 0.7997  & 0.8499  \\
    NCC   & \multicolumn{1}{c}{0.6701} & \multicolumn{1}{c}{0.7184} & \multicolumn{1}{c}{\textbf{0.7582}} & \multicolumn{1}{c}{0.4952} & \multicolumn{1}{c}{\textcolor[rgb]{ 1,  0,  0}{\textbf{0.8159}}} & \multicolumn{1}{c}{\textcolor[rgb]{ 0,  0,  1}{\textbf{0.7687}}} & \multicolumn{1}{c}{0.7323} & \multicolumn{1}{c}{0.2966} & 0.7341  & 0.7380  \\
    SCN   & \multicolumn{1}{c}{0.6659} & \multicolumn{1}{c}{0.1695} & \multicolumn{1}{c}{0.3354} & \multicolumn{1}{c}{\textbf{0.6967}} & \multicolumn{1}{c}{\textcolor[rgb]{ 1,  0,  0}{\textbf{0.9233}}} & \multicolumn{1}{c}{0.5804} & \multicolumn{1}{c}{0.6507} & \multicolumn{1}{c}{0.0119} & 0.5806  & \textcolor[rgb]{ 0,  0,  1}{\textbf{0.8150}} \\
    MN    & \multicolumn{1}{c}{\textcolor[rgb]{ 1,  0,  0}{\textbf{0.7464}}} & \multicolumn{1}{c}{0.5927} & \multicolumn{1}{c}{0.5752} & \multicolumn{1}{c}{0.0468} & \multicolumn{1}{c}{0.5135} & \multicolumn{1}{c}{\textbf{0.7250}} & \multicolumn{1}{c}{\textcolor[rgb]{ 0,  0,  1}{\textbf{0.7383}}} & \multicolumn{1}{c}{0.6624} & 0.6582  & 0.6490  \\
    HFN   & \multicolumn{1}{c}{0.8454} & \multicolumn{1}{c}{0.8628} & \multicolumn{1}{c}{\textcolor[rgb]{ 0,  0,  1}{\textbf{0.8923}}} & \multicolumn{1}{c}{\textcolor[rgb]{ 1,  0,  0}{\textbf{0.9250}}} & \multicolumn{1}{c}{0.8691} & \multicolumn{1}{c}{0.8642} & \multicolumn{1}{c}{0.8730} & \multicolumn{1}{c}{0.8214} & 0.8794  & \textbf{0.8875} \\
    IN    & \multicolumn{1}{c}{0.7437} & \multicolumn{1}{c}{\textcolor[rgb]{ 0,  0,  1}{\textbf{0.8003}}} & \multicolumn{1}{c}{0.6901} & \multicolumn{1}{c}{0.4324} & \multicolumn{1}{c}{0.7556} & \multicolumn{1}{c}{\textbf{0.7878}} & \multicolumn{1}{c}{\textcolor[rgb]{ 1,  0,  0}{\textbf{0.8006}}} & \multicolumn{1}{c}{0.5677} & 0.7138  & 0.7703  \\
    QN    & \multicolumn{1}{c}{0.8503} & \multicolumn{1}{c}{0.7089} & \multicolumn{1}{c}{0.7508} & \multicolumn{1}{c}{0.8536} & \multicolumn{1}{c}{\textcolor[rgb]{ 0,  0,  1}{\textbf{0.8714}}} & \multicolumn{1}{c}{0.7991} & \multicolumn{1}{c}{\textbf{0.8573}} & \multicolumn{1}{c}{0.7732} & 0.7357  & \textcolor[rgb]{ 1,  0,  0}{\textbf{0.8729}} \\
    GB    & \multicolumn{1}{c}{0.7969} & \multicolumn{1}{c}{0.8464} & \multicolumn{1}{c}{0.8280} & \multicolumn{1}{c}{0.8357} & \multicolumn{1}{c}{0.8145} & \multicolumn{1}{c}{\textcolor[rgb]{ 1,  0,  0}{\textbf{0.9046}}} & \multicolumn{1}{c}{\textcolor[rgb]{ 0,  0,  1}{\textbf{0.8628}}} & \multicolumn{1}{c}{0.7595} & 0.7852  & \textbf{0.8614} \\
    ID    & \multicolumn{1}{c}{0.5901} & \multicolumn{1}{c}{0.3381} & \multicolumn{1}{c}{\textbf{0.6442}} & \multicolumn{1}{c}{0.2487} & \multicolumn{1}{c}{\textcolor[rgb]{ 0,  0,  1}{\textbf{0.7494}}} & \multicolumn{1}{c}{0.0690} & \multicolumn{1}{c}{0.6118} & \multicolumn{1}{c}{0.6403} & 0.5854  & \textcolor[rgb]{ 1,  0,  0}{\textbf{0.8752}} \\
    JPEG  & \multicolumn{1}{c}{0.8427} & \multicolumn{1}{c}{0.8369} & \multicolumn{1}{c}{0.7929} & \multicolumn{1}{c}{\textcolor[rgb]{ 1,  0,  0}{\textbf{0.9122}}} & \multicolumn{1}{c}{0.8343} & \multicolumn{1}{c}{\textcolor[rgb]{ 0,  0,  1}{\textbf{0.9115}}} & \multicolumn{1}{c}{0.8775} & \multicolumn{1}{c}{0.8474} & 0.6507  & \textbf{0.8952} \\
    J2K   & \multicolumn{1}{c}{0.8890} & \multicolumn{1}{c}{0.7895} & \multicolumn{1}{c}{0.8536} & \multicolumn{1}{c}{\textbf{0.8983}} & \multicolumn{1}{c}{0.8583} & \multicolumn{1}{c}{\textcolor[rgb]{ 0,  0,  1}{\textbf{0.9194}}} & \multicolumn{1}{c}{0.8813} & \multicolumn{1}{c}{0.8507} & 0.8242  & \textcolor[rgb]{ 1,  0,  0}{\textbf{0.9326}} \\
    JTE   & \multicolumn{1}{c}{0.0727} & \multicolumn{1}{c}{0.0491} & \multicolumn{1}{c}{0.2287} & \multicolumn{1}{c}{0.0912} & \multicolumn{1}{c}{\textbf{0.3628}} & \multicolumn{1}{c}{\textcolor[rgb]{ 1,  0,  0}{\textbf{0.7085}}} & \multicolumn{1}{c}{0.3214} & \multicolumn{1}{c}{0.0343} & 0.2019  & \textcolor[rgb]{ 0,  0,  1}{\textbf{0.4233}} \\
    J2KTE & \multicolumn{1}{c}{0.5250} & \multicolumn{1}{c}{0.4061} & \multicolumn{1}{c}{0.1129} & \multicolumn{1}{c}{\textcolor[rgb]{ 0,  0,  1}{\textbf{0.6106}}} & \multicolumn{1}{c}{\textbf{0.6085}} & \multicolumn{1}{c}{0.3651} & \multicolumn{1}{c}{\textcolor[rgb]{ 1,  0,  0}{\textbf{0.6107}}} & \multicolumn{1}{c}{0.0096} & 0.0962  & 0.5262  \\
    NEPN  & \multicolumn{1}{c}{\textbf{0.0687}} & \multicolumn{1}{c}{0.0479} & \multicolumn{1}{c}{0.0100} & \multicolumn{1}{c}{0.0522} & \multicolumn{1}{c}{\textcolor[rgb]{ 0,  0,  1}{\textbf{0.0809}}} & \multicolumn{1}{c}{\textcolor[rgb]{ 1,  0,  0}{\textbf{0.3714}}} & \multicolumn{1}{c}{0.0073} & \multicolumn{1}{c}{0.0621} & 0.0126  & 0.0165  \\
    LBD   & \multicolumn{1}{c}{0.1305} & \multicolumn{1}{c}{\textbf{0.2473}} & \multicolumn{1}{c}{0.1778} & \multicolumn{1}{c}{0.1374} & \multicolumn{1}{c}{0.1317} & \multicolumn{1}{c}{\textcolor[rgb]{ 1,  0,  0}{\textbf{0.2912}}} & \multicolumn{1}{c}{0.0328} & \multicolumn{1}{c}{0.0901} & \textcolor[rgb]{ 0,  0,  1}{\textbf{0.2882}} & 0.0889  \\
    MS    & \multicolumn{1}{c}{0.1627} & \multicolumn{1}{c}{\textcolor[rgb]{ 0,  0,  1}{\textbf{0.3060}}} & \multicolumn{1}{c}{\textbf{0.2784}} & \multicolumn{1}{c}{\textcolor[rgb]{ 1,  0,  0}{\textbf{0.3406}}} & \multicolumn{1}{c}{0.1843} & \multicolumn{1}{c}{0.0987} & \multicolumn{1}{c}{0.0649} & \multicolumn{1}{c}{0.0956} & 0.1191  & 0.1812  \\
    CC    & \multicolumn{1}{c}{0.0172} & \multicolumn{1}{c}{0.2067} & \multicolumn{1}{c}{0.0715} & \multicolumn{1}{c}{0.1994} & \multicolumn{1}{c}{0.0144} & \multicolumn{1}{c}{0.1447} & \multicolumn{1}{c}{0.1369} & \multicolumn{1}{c}{\textcolor[rgb]{ 1,  0,  0}{\textbf{0.4623}}} & \textbf{0.2501} & \textcolor[rgb]{ 0,  0,  1}{\textbf{0.2840}} \\
    CCS   & \multicolumn{1}{c}{0.2462} & \multicolumn{1}{c}{\textbf{0.3691}} & \multicolumn{1}{c}{0.2682} & \multicolumn{1}{c}{0.3017} & \multicolumn{1}{c}{0.1654} & \multicolumn{1}{c}{0.0700} & \multicolumn{1}{c}{0.1316} & \multicolumn{1}{c}{\textcolor[rgb]{ 0,  0,  1}{\textbf{0.3781}}} & 0.1349  & \textcolor[rgb]{ 1,  0,  0}{\textbf{0.5557}} \\
    MGN   & \multicolumn{1}{c}{0.6934} & \multicolumn{1}{c}{\textcolor[rgb]{ 0,  0,  1}{\textbf{0.7902}}} & \multicolumn{1}{c}{0.7322} & \multicolumn{1}{c}{0.6960} & \multicolumn{1}{c}{0.6936} & \multicolumn{1}{c}{\textbf{0.7882}} & \multicolumn{1}{c}{0.7406} & \multicolumn{1}{c}{0.3958} & \textcolor[rgb]{ 1,  0,  0}{\textbf{0.8026}} & 0.7590  \\
    CN    & \multicolumn{1}{c}{0.1914} & \multicolumn{1}{c}{0.1523} & \multicolumn{1}{c}{0.1475} & \multicolumn{1}{c}{0.0180} & \multicolumn{1}{c}{\textcolor[rgb]{ 1,  0,  0}{\textbf{0.3941}}} & \multicolumn{1}{c}{\textcolor[rgb]{ 0,  0,  1}{\textbf{0.3909}}} & \multicolumn{1}{c}{0.2242} & \multicolumn{1}{c}{0.1370} & 0.3119  & \textbf{0.3181} \\
    LCN   & \multicolumn{1}{c}{0.8025} & \multicolumn{1}{c}{0.6399} & \multicolumn{1}{c}{0.6369} & \multicolumn{1}{c}{0.2356} & \multicolumn{1}{c}{0.8287} & \multicolumn{1}{c}{\textcolor[rgb]{ 1,  0,  0}{\textbf{0.8513}}} & \multicolumn{1}{c}{\textbf{0.8307}} & \multicolumn{1}{c}{0.3429} & 0.7700  & \textcolor[rgb]{ 0,  0,  1}{\textbf{0.8458}} \\
    ICQ   & \multicolumn{1}{c}{0.7827} & \multicolumn{1}{c}{\textcolor[rgb]{ 0,  0,  1}{\textbf{0.8733}}} & \multicolumn{1}{c}{\textbf{0.8119}} & \multicolumn{1}{c}{\textcolor[rgb]{ 1,  0,  0}{\textbf{0.8969}}} & \multicolumn{1}{c}{0.7496} & \multicolumn{1}{c}{0.7562} & \multicolumn{1}{c}{0.7890} & \multicolumn{1}{c}{0.7556} & 0.0911  & 0.8043  \\
    CA    & \multicolumn{1}{c}{0.5620} & \multicolumn{1}{c}{0.6250} & \multicolumn{1}{c}{0.6756} & \multicolumn{1}{c}{\textbf{0.6953}} & \multicolumn{1}{c}{0.6793} & \multicolumn{1}{c}{\textcolor[rgb]{ 0,  0,  1}{\textbf{0.6998}}} & \multicolumn{1}{c}{0.6339} & \multicolumn{1}{c}{0.5816} & 0.5206  & \textcolor[rgb]{ 1,  0,  0}{\textbf{0.7177}} \\
    SSR & \multicolumn{1}{c}{0.8340} & \multicolumn{1}{c}{0.7857} & \multicolumn{1}{c}{0.8229} & \multicolumn{1}{c}{\textbf{0.8580}} & \multicolumn{1}{c}{\textcolor[rgb]{ 0,  0,  1}{\textbf{0.8643}}} & \multicolumn{1}{c}{0.7610} & \multicolumn{1}{c}{0.8284} & \multicolumn{1}{c}{0.8251} & 0.7245  & \textcolor[rgb]{ 1,  0,  0}{\textbf{0.9196}} \\
    \bottomrule
    \bottomrule
    \end{tabular}%
  \label{tab:distortion_type}%
\end{table*}%

\subsection{Quality Calculation}
\label{sec:index_calculation}

The proposed quality index is computed by measuring the distance between the MVG model fitted using the features of the testing image and the \revise{benchmark} MVG model fitted using the features from a high-quality image set. All the deep features are utilized in the statistical data analysis model for the testing image without removing low contrast features as suggested in~\cite{mittal2012making}. 
\revise{Specifically, the quality score of the test image is calculated by the Mahalanobis distance between two MVGs as suggested in~\cite{zhang2015feature, mittal2012making, liu2019unsupervised}. The covariance matrix is defined as the average of the two covariances. Therefore, the quality score is calculated as:}
\begin{equation}
\begin{aligned}
S(\overrightarrow{\mu_d}, \overrightarrow{\mu_r}, &\sum\nolimits_d, \sum\nolimits_r) =  \\
&\sqrt{\left(\left(\overrightarrow{\mu_d}-\overrightarrow{\mu_r}\right)^{\rm T}\left(\frac{\sum_d + \sum_r}{2}\right)^{-1} \left(\overrightarrow{\mu_d}-\overrightarrow{\mu_r}\right)\right)},
\end{aligned}
\end{equation}
\revise{where $\{\overrightarrow{\mu_d}, \sum_d\}$ and $\{\overrightarrow{\mu_r}$, $\sum_r\}$ denote the mean vectors and covariance matrices of the estimated MVG models of the distorted image and the high-quality image set, respectively.}

\section{Experiment}
\label{sec:experiment}
In this section, we first describe the basic experimental protocol to ensure a fair comparison with existing BIQA methods. 
We then evaluate the performance of the proposed model compared with classicial and state-of-the-art models. 
Finally, we conduct extensive ablation studies to further evaluate the effectiveness of the custom modules and analyze the limitation of the proposed MDFS.

\begin{figure*}[htbp]
\centering
\captionsetup[subfloat]{labelsep=none,format=plain,labelformat=empty}
    \subfloat[NIQE]{
        \includegraphics[width=0.23\textwidth]{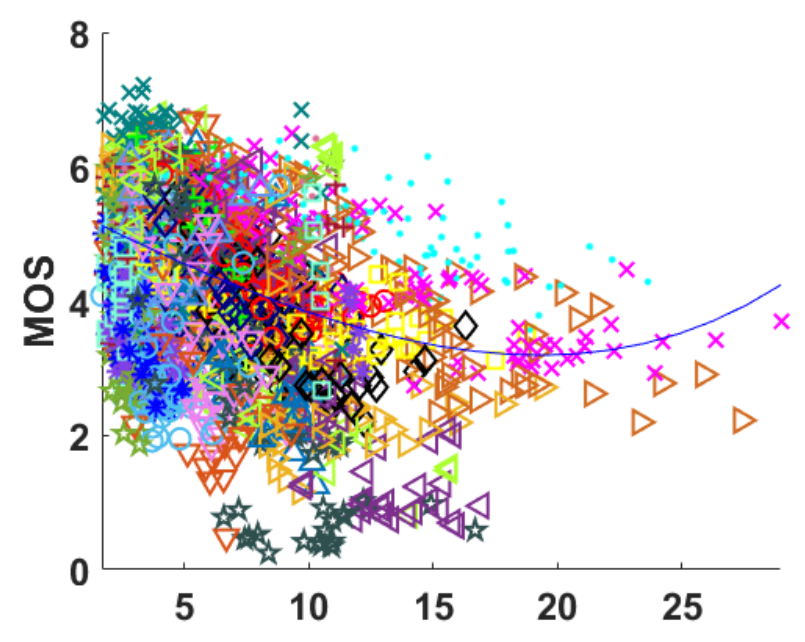}
}
    \subfloat[QAC]{
        \includegraphics[width=0.23\textwidth]{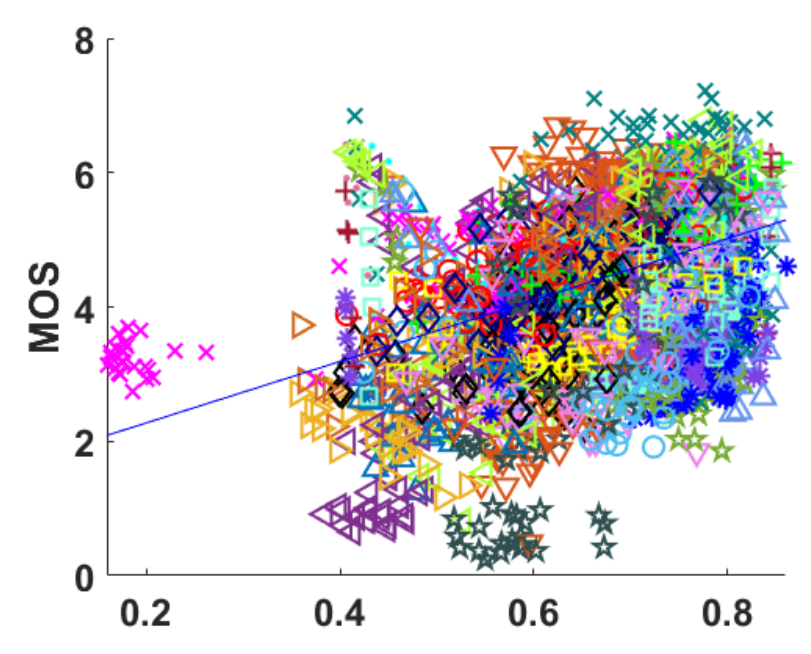}
}
    \subfloat[PIQE]{
        \includegraphics[width=0.23\textwidth]{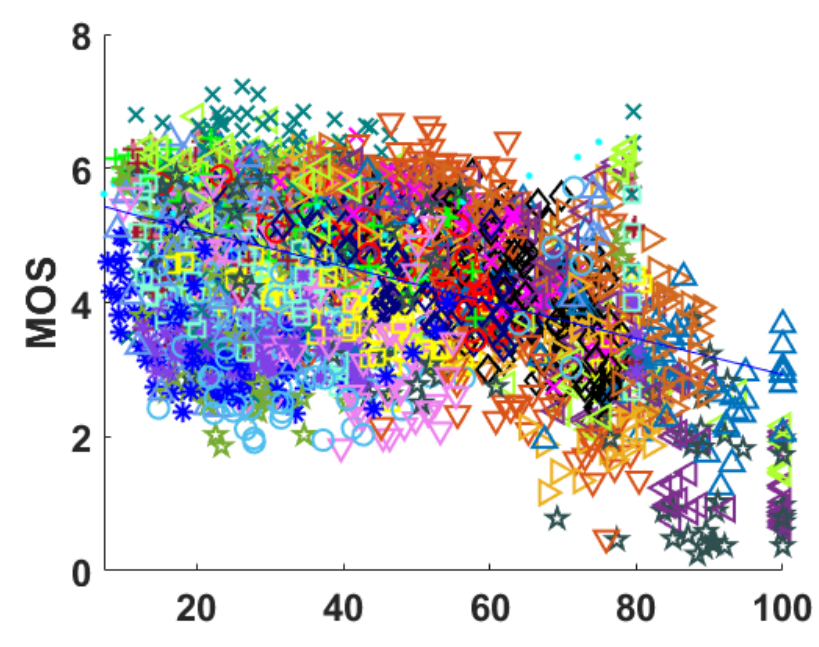}
}
    \subfloat[LPSI]{
        \includegraphics[width=0.23\textwidth]{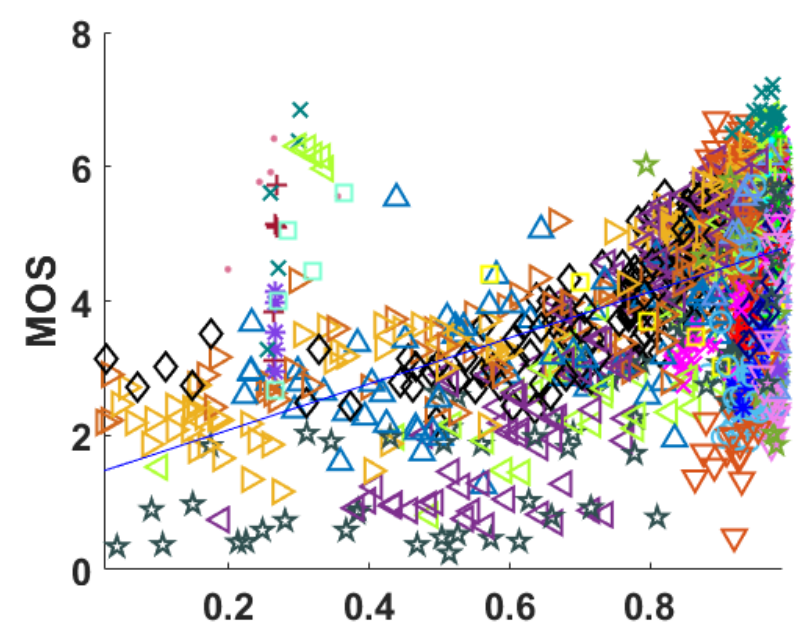}
} 
\\ 
    \subfloat[ILNIQE]{
        \includegraphics[width=0.23\textwidth]{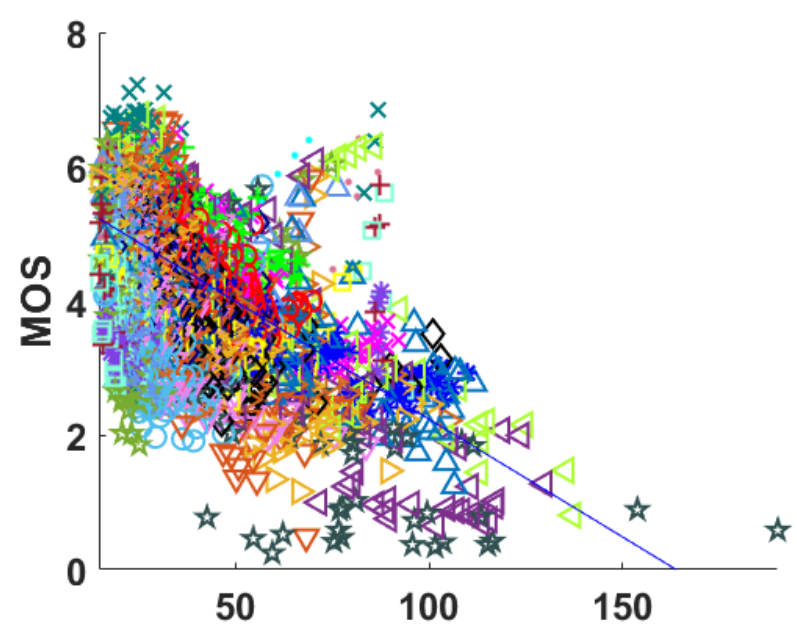}
}
    \subfloat[dipIQ]{
        \includegraphics[width=0.23\textwidth]{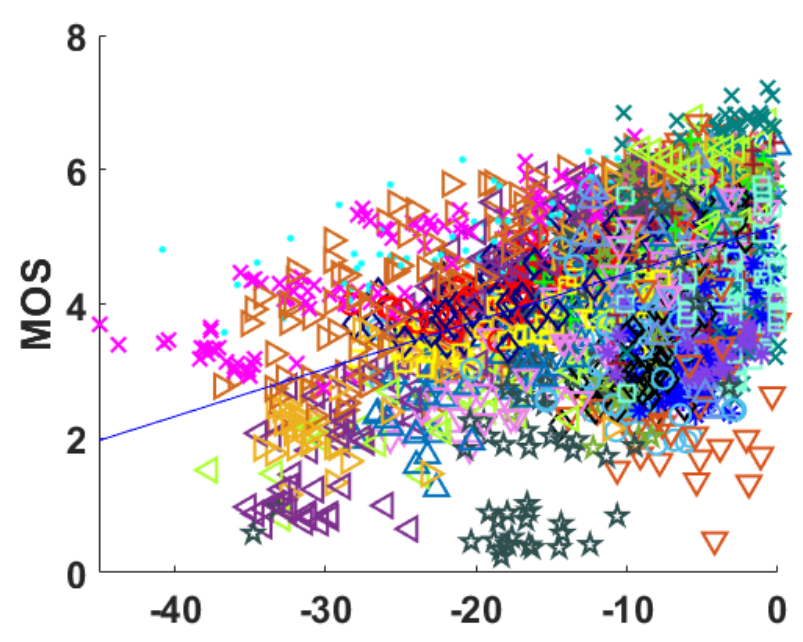}
} 
    \subfloat[SNP-NIQE]{
        \includegraphics[width=0.23\textwidth]{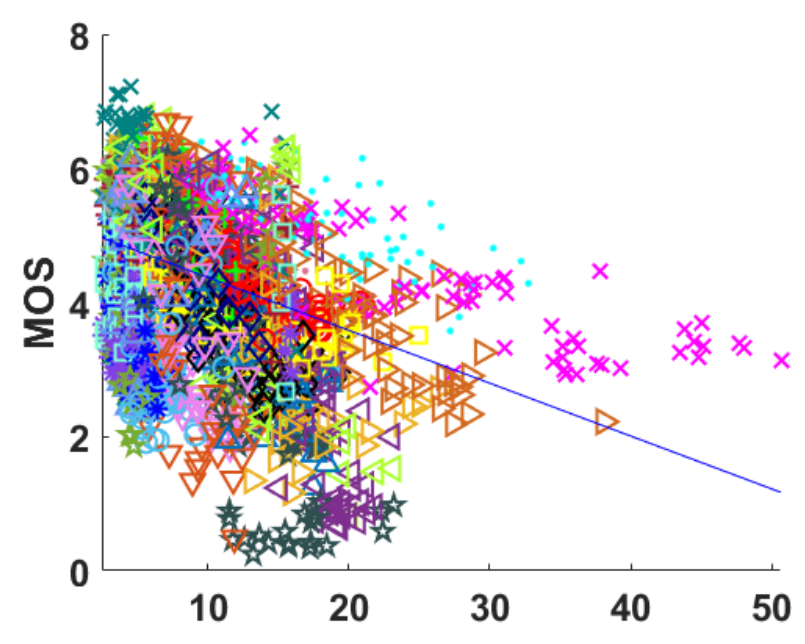}
}
    \subfloat[NPQI]{
        \includegraphics[width=0.23\textwidth]{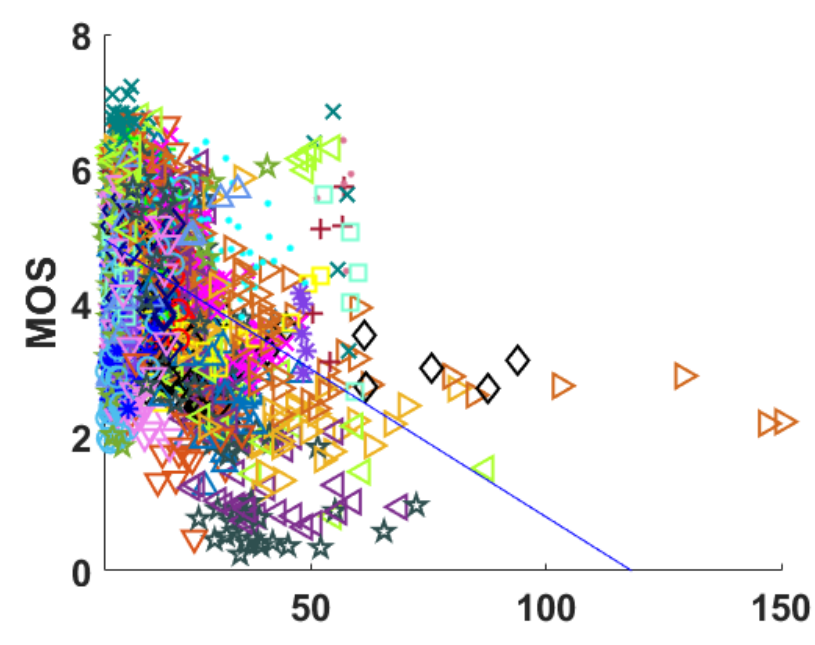}
}
\\
    \subfloat[ContentSep]{
        \includegraphics[width=0.23\textwidth]{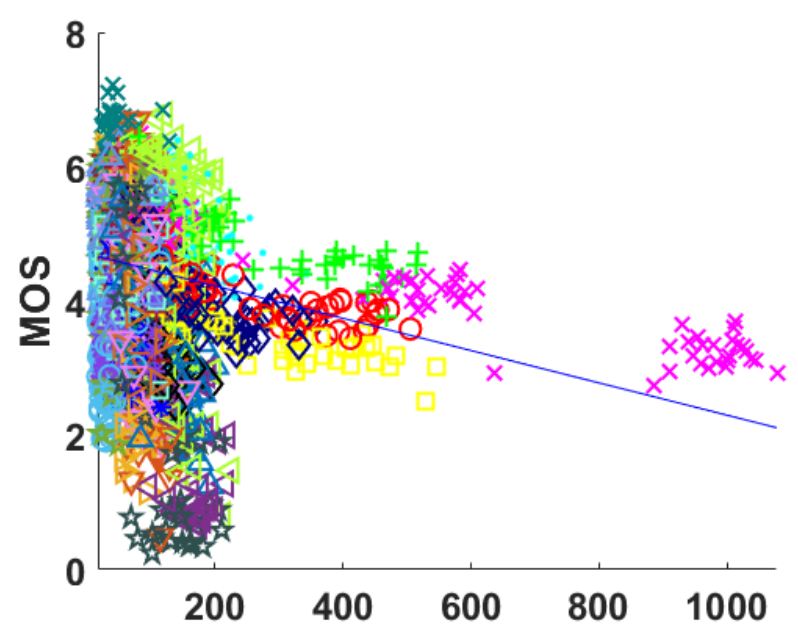}
}
   \subfloat[MDFS (Ours)]{
        \includegraphics[width=0.23\textwidth]{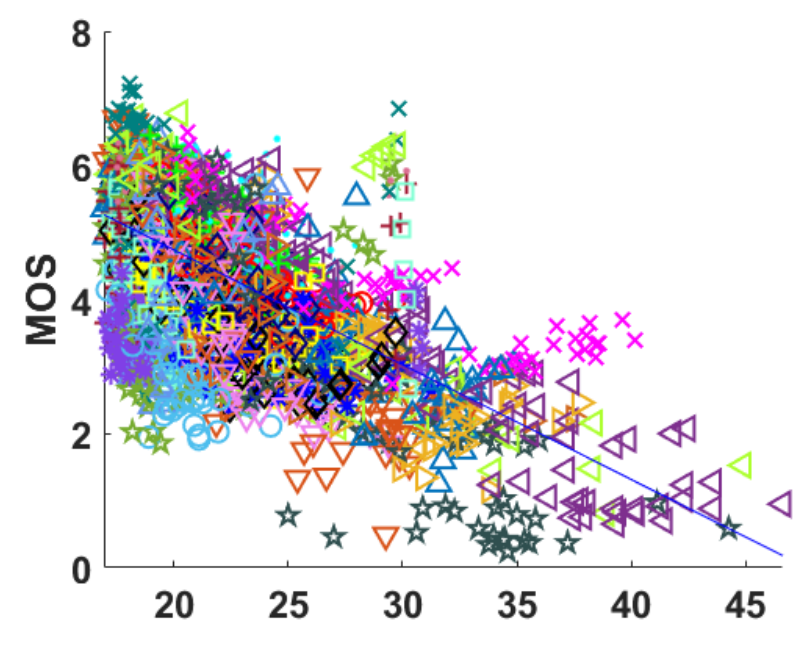}
}
    \subfloat[Legend]{
        \includegraphics[width=0.46\textwidth]{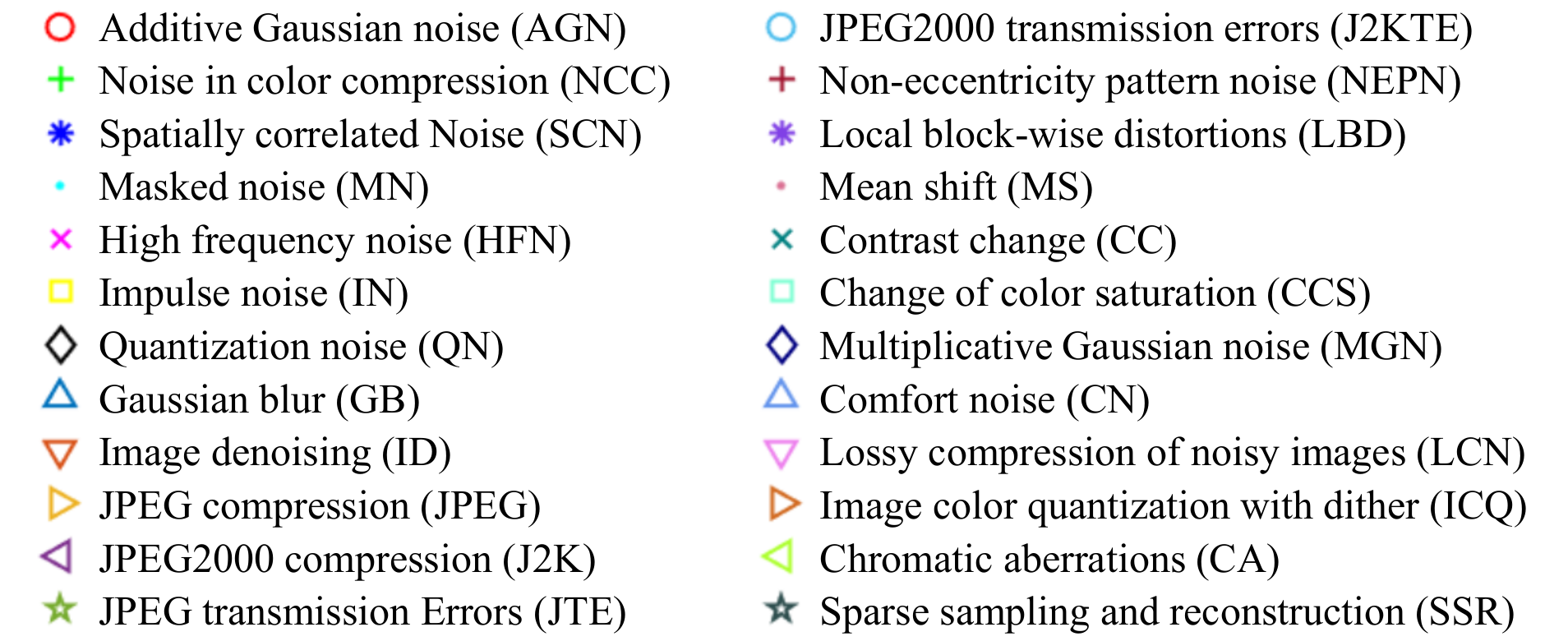}
}
\caption{Scatter plots of the mean opinion scores (MOS) versus the objective scores computed by the IQA models: (a) NIQE; (b) QAC; (c) PIQE; (d) LPSI; (e) ILNIQE; (f) dipIQ; (g) SNP-NIQE; (h) NPQI; (i) ContentSep, and (j) MDFS, respectively.
}
\label{fig:scatter_plots}
\end{figure*}

\subsection{Experiment Protocol}
\label{sec:setting}

\subsubsection{Training Dataset}
In this work, we collect a training dataset including 500 high-quality natural images with various image sizes and content types from the DIV2K~\cite{Agustsson2017} dataset, which is non-overlapped with the testing dataset. However, we study the impact of different training datasets on our proposed method in the subsequent ablation experiment section.

\subsubsection{Testing Datasets}
To comprehensively evaluate the performance of the proposed model, various IQA datasets have been used to conduct extensive experiments, including CLIVE~\cite{ghadiyaram2015massive}, CID 2013~\cite{virtanen2014cid2013}, KonIQ~\cite{hosu2020koniq}, SPAQ~\cite{fang2020perceptual}, LIVE~\cite{sheikh2006statistical}, CSIQ~\cite{larson2010most}, TID2013~\cite{ponomarenko2015image}, KADID~\cite{lin2019kadid},  MDLIVE~\cite{6489321}, and MDIVL~\cite{corchs2014noisy}.
Specifically, the above datasets can be classified into two categories according to the degradation methods, where the first four and last six datasets belong to the realistic distortion-based datasets and synthetic distortion-based datasets, respectively.

\subsubsection{Evaluation Criteria}
Following~\cite{ni2017esim, zhang2015feature}, the predicted objective scores are first mapped to the subjective scores by a nonlinear regression function as:
\begin{equation}
Q_i(x_i) = \gamma_1\left(\frac{1}{2}-\frac{1}{1+e^{\gamma_2(x_i-\gamma_3)}}\right) + \gamma_4x_i + \gamma_5,
\label{map_equ}
\end{equation}
where $x_i$ and $Q_i$ are the predicted score and corresponding mapped score, respectively. The five parameters $\gamma_1, \gamma_2, \gamma_3, \gamma_4$ and $\gamma_5$ are determined by fitting the regression model.
To evaluate the effectiveness of the proposed MDFS and comparison models, we employ four common evaluation criteria: Spearman rank order correlation coefficient (SROCC), Kendall rank order correlation coefficient (KROCC), Pearson linear correlation coefficient (PLCC), and Root-mean-square error (RMSE). 
It should be noted that a better IQA method should yield higher values for SROCC, KRCC, PLCC, and lower values for RMSE.

\begin{table*}[t]
  \centering
  \setlength{\tabcolsep}{3pt}
  \caption{
  Performance comparisons of the original/retrained OU-BIQA models on four target-specific IQA datasets. 
 }
        \begin{tabular}{cl|cc|cc|cc|cc|cc|cc}
    \toprule
    \toprule
    \multicolumn{2}{c|}{} & \multicolumn{6}{c|}{SROCC $\uparrow$}   & \multicolumn
    {6}{c}{KROCC $\uparrow$} \\
    \cmidrule{3-14}
     \multicolumn{2}{c|}{Datasets} & \multicolumn{2}{c|}{NIQE} & \multicolumn{2}{c|}{ILNIQE} & \multicolumn{2}{c|}{MDFS (ours)} & \multicolumn{2}{c|}{NIQE} & \multicolumn{2}{c|}{ILNIQE} & \multicolumn{2}{c}{MDFS (ours)} \\
    \cmidrule{3-14}
     & & Original & Retrained & Original & Retrained & Original & Retrained & Original & Retrained & Original & Retrained & Original & Retrained \\
    \midrule
    \multirow{2}{*}{UWI} & \multicolumn{1}{c|}{SAUD} & 0.0616  & 0.1058  & 0.2818  & 0.2809  & 0.3353  & \textcolor[rgb]{ 1,  0,  0}{\textbf{0.4431}} & 0.0417  & 0.0717  & 0.1920  & 0.1920  & 0.2285  & \textcolor[rgb]{ 1,  0,  0}{\textbf{0.3098}} \\
          & \multicolumn{1}{c|}{UWIQA} & 0.4484  & 0.4803  & 0.4718  & 0.3440  & 0.3020  & \textcolor[rgb]{ 1,  0,  0}{\textbf{0.6014}} & 0.3348  & 0.3576  & 0.3490  & 0.2473  & 0.2202  & \textcolor[rgb]{ 1,  0,  0}{\textbf{0.4594}} \\
    \midrule
    \multirow{2}{*}{AIGC} & \multicolumn{1}{c|}{AGIQA} & 0.5338  & -     & 0.5943  & -     & \textcolor[rgb]{ 1,  0,  0}{\textbf{0.6724}} & -     & 0.3651  & -     & 0.4194  & -     & \textcolor[rgb]{ 1,  0,  0}{\textbf{0.4816}} & - \\
          & \multicolumn{1}{c|}{AIGCIQA} & 0.5062  & -     & 0.5692  & -     & \textcolor[rgb]{ 1,  0,  0}{\textbf{0.6992}} & -     & 0.3422  & -     & 0.3852  & -     & \textcolor[rgb]{ 1,  0,  0}{\textbf{0.4887}} & - \\
    \midrule
    Bird  & MMQA-Birds & 0.3098  & 0.0989  & 0.1226  & 0.2367  & 0.1883  & \textcolor[rgb]{ 1,  0,  0}{\textbf{0.4974}} & 0.2119  & 0.0674  & 0.0842  & 0.1627  & 0.1273  & \textcolor[rgb]{ 1,  0,  0}{\textbf{0.3473}} \\
    \midrule
    Face  & GFIQA-20k & 0.5011  & 0.6655  & 0.7142  & 0.7656  & 0.8331  & \textcolor[rgb]{ 1,  0,  0}{\textbf{0.8359}} & 0.3495  & 0.4793  & 0.5184  & 0.5720  & 0.6410  & \textcolor[rgb]{ 1,  0,  0}{\textbf{0.6479}} \\
    \midrule
    & & \multicolumn{6}{c|}{PLCC $\uparrow$}     & \multicolumn{6}{c}{RMSE $\downarrow$} \\
    \cmidrule{3-14}
    \multicolumn{2}{c|}{Datasets} & \multicolumn{2}{c|}{NIQE} & \multicolumn{2}{c|}{ILNIQE} & \multicolumn{2}{c|}{MDFS (ours)} & \multicolumn{2}{c|}{NIQE} & \multicolumn{2}{c|}{ILNIQE} & \multicolumn{2}{c}{MDFS (ours)} \\
    \cmidrule{3-14}
    & & Original & Retrained & Original & Retrained & Original & Retrained & Original & Retrained & Original & Retrained & Original & Retrained \\
    \midrule
    \multirow{2}{*}{UWI} & \multicolumn{1}{c|}{SAUD} & 0.0876  & 0.1569  & 0.3141  & 0.3327  & 0.3663  & \textcolor[rgb]{ 1,  0,  0}{\textbf{0.5334}} & 1.5500  & 1.5367  & 1.4772  & 1.4674  & 1.4479  & \textcolor[rgb]{ 1,  0,  0}{\textbf{1.3162}} \\
          & \multicolumn{1}{c|}{UWIQA} & 0.4451  & 0.3373  & 0.4062  & 0.3368  & 0.4242  & \textcolor[rgb]{ 1,  0,  0}{\textbf{0.5588}} & 0.1364  & 0.1434  & 0.1392  & 0.1434  & 0.1955  & \textcolor[rgb]{ 1,  0,  0}{\textbf{0.1263}} \\
    \midrule
    \multirow{2}{*}{AIGC} & \multicolumn{1}{c|}{AGIQA} & 0.5391  & -     & 0.6229  & -     & \textcolor[rgb]{ 1,  0,  0}{\textbf{0.6762}} & -     & 0.8377  & -     & 0.7805  & -     & \textcolor[rgb]{ 1,  0,  0}{\textbf{0.7350}} & - \\
          & \multicolumn{1}{c|}{AIGCIQA} & 0.5219  & -     & 0.5641  & -     & \textcolor[rgb]{ 1,  0,  0}{\textbf{0.7047}} & -     & 7.9454  & -     & 7.6914  & -     & \textcolor[rgb]{ 1,  0,  0}{\textbf{6.6087}} & - \\
    \midrule
    Bird  & MMQA-Birds & 0.3125  & 0.1167  & 0.1492  & 0.2476  & 0.2260  & \textcolor[rgb]{ 1,  0,  0}{\textbf{0.4531}} & 1.5988  & 1.6716  & 1.6642  & 1.6306  & 1.6395  & \textcolor[rgb]{ 1,  0,  0}{\textbf{1.5004}} \\
    \midrule
    Face  & GFIQA-20k & 0.5006  & 0.6639  & 0.7022  & 0.6910  & \textcolor[rgb]{ 1,  0,  0}{\textbf{0.8070}} & 0.7948  & 0.1580  & 0.1341  & 0.1299  & 0.1297  & \textcolor[rgb]{ 1,  0,  0}{\textbf{0.1078}} & 0.1089  \\
    \bottomrule
    \bottomrule
    \end{tabular}%
  \label{tab:generalizability}%
\end{table*}%

\subsection{Comparisons with State-of-the-Arts}
\label{sec:compraisions}
In this subsection, we first compare the proposed MDFS with nine classical and state-of-the-art OU-BIQA methods, including NIQE~\cite{mittal2012making}, \wh{QAC~}\cite{xue2013learning}, \wh{PIQE~\cite{venkatanath2015blind}, LPSI~\cite{wu2015highly}}, ILNIQE~\cite{zhang2015feature}, dipIQ~\cite{ma2017dipiq}, SNP-NIQE~\cite{liu2019unsupervised}, NPQI~\cite{liu2020blind}, and ContentSep~\cite{babu2023no}. 
Our focus is on demonstrating the advantages of our proposed MDFS across multiple datasets and diverse types of distortions. Detailed findings are presented below.

\subsubsection{Performance on Multiple IQA datasets}
Table~\ref{tab:overall} presents the performance comparison of various BIQA methods on ten datasets. 
Notably, the top three performers for each measurement criterion (\textit{i.e.}, PLCC, SROCC, KROCC, and RMSE) are highlighted, with the first-ranked, second-ranked, and third-ranked IQA models emphasized in bold red, blue, and black, respectively. 
From the results, we can observe that the proposed MDFS significantly outperforms state-of-the-art OU-BIQA models on TID2013, KADID, MDLIVE, KonIQ, CID2013, and SPAQ datasets in terms of the PLCC, SROCC, KRCC, and RMSE. 
Additionally, on the LIVE, CSIQ, and CLIVE datasets, MDFS ranks top third in overall performance and is almost comparable with the state-of-the-art models.

To evaluate the overall performance across multiple datasets, two average measurements are applied as suggested in~\cite{ni2018gabor},
\begin{equation}
\overline{c} = \sum\nolimits_{i=1}^{N} (c_i \cdot w_i) \bigg/ \sum\nolimits_{i=1}^{N} w_i,
\end{equation}
where $N$ denotes the number of datasets, $c_i$ and $w_i$ indicate the value of the measurement criteria and corresponding weight on the $i$-th dataset. Herein, we first set all the weights to 1 to obtain the results of the \textit{Direct Average} ($AVG_D$). Afterward, $w_i$ is set to the number of the distorted images in the $i$-th dataset to obtain the results of the \textit{Weighted Average} ($AVG_W$). 
Due to the large difference in the range of quality scores obtained by various IQA models, the average RMSE is not applicable for fairness. 
From Table~\ref{tab:overall}, the proposed MDFS is superior to all existing BIQA models in terms of both direct average and weighted average comparisons, which demonstrates the generality of the proposed model.

\subsubsection{Performance on Diverse Distortion Types}
In this section, we conduct a detailed and comprehensive experiment to evaluate the performance of the proposed MDFS model across various types of distortions using the TID2013 dataset. From Table~\ref{tab:distortion_type}, one can observe that our model yields the most top-one and top-three performances compared with other IQA models in terms of SROCC. Specifically, in the comparisons in terms of SROCC, the proposed MDFS stands in the top three 14 times and ranks first 6 times, which is the same as the dipIQ model. 
The following are the ILNIQE, and LPSI, which ranked the top three 10 times and 9 times respectively, and ranked the top one 3 times and 4 times, respectively.

To visually compare the performance of BIQA models, Figure~\ref{fig:scatter_plots} provides scatter plots of subjective scores (\textit{i.e.}, MOS) and objective scores, where all the objective scores are generated by the IQA models and further mapped using Equ.~(\ref{map_equ}). 
The blue lines in each sub-figure represents the fitted line obtained from Equ.~(\ref{map_equ}), indicating the \wh{``mean''} value of the performance. 
For each distortion type, the predicted quality score (along the horizontal axis) is expected to be close to the MOS value (along the vertical axis). 
Therefore, the closer the points representing a specific type of distortion are clustered around this line, the better the algorithm performs on that type of distortion. 
For example, for the JPEG2000 compression (J2K) distortion type, the predicted quality scores are closer to the corresponding fitted blue line in the scatter plot of the proposed MDFS, while predicted quality scores of dipIQ and LPSI are far away from the corresponding fitted blue line.
This demonstrates that the quality scores predicted by the proposed MDFS are more consistent with the HVS regarding the J2K distortion, aligning with the results shown in Table~\ref{tab:distortion_type}. 
Beyond the localized examination of scatter plots, an overarching perspective facilitates the observation of the interplay between various distortion types and the reference blue line. 
For instance, it is obvious that the proposed MDFS framework exhibits a closer fit to the blue line on various distortion types than other BIQA models.
This coherent alignment substantiates the comprehensive superiority of the performance rendered by the proposed model, which is consistent with the results shown in Table~\ref{tab:overall}.

\begin{table*}[t]
  \centering
  \setlength{\tabcolsep}{2.5pt}
  \caption{Performance comparisons between our proposed OU-BIQA and different OA-BIQA methods on the public datasets. 
  }
    \begin{tabular}{c|ccccccc|ccccccc}
    \toprule
    \toprule
    \multicolumn{1}{r}{} &       & PaQ2PiQ & HyperIQA & MANIQA & VCRNet & MUSIQ & \multicolumn{1}{c}{MDFS (Ours)} &       & PaQ2PiQ & HyperIQA & MANIQA & VCRNet & MUSIQ & MDFS (Ours) \\
    \midrule
    LIVE  & \multirow{12}[6]{*}{\begin{sideways}SROCC\end{sideways}} & 0.4794  & \textbf{0.7551} & \textcolor[rgb]{ 0,  0,  1}{\textbf{0.7793}} & -     & 0.7335  & \textcolor[rgb]{ 1,  0,  0}{\textbf{0.9361}} & \multirow{12}[6]{*}{\begin{sideways}KROCC\end{sideways}} & 0.3557  & \textbf{0.5515} & \textcolor[rgb]{ 0,  0,  1}{\textbf{0.5736}} & -     & 0.5422  & \textcolor[rgb]{ 1,  0,  0}{\textbf{0.7709}} \\
    CSIQ  &       & 0.5643  & 0.5814  & \textbf{0.6624} & \textcolor[rgb]{ 0,  0,  1}{\textbf{0.6806}} & 0.5878  & \textcolor[rgb]{ 1,  0,  0}{\textbf{0.7774}} &       & 0.4000  & 0.4004  & \textbf{0.4679} & \textcolor[rgb]{ 0,  0,  1}{\textbf{0.5031}} & 0.4129  & \textcolor[rgb]{ 1,  0,  0}{\textbf{0.5823}} \\
    TID2013 &       & 0.4011  & 0.3839  & 0.4510  & \textcolor[rgb]{ 0,  0,  1}{\textbf{0.5116}} & \textbf{0.4738} & \textcolor[rgb]{ 1,  0,  0}{\textbf{0.5363}} &       & 0.2838  & 0.2606  & 0.3175  & \textcolor[rgb]{ 0,  0,  1}{\textbf{0.3667}} & \textbf{0.3314} & \textcolor[rgb]{ 1,  0,  0}{\textbf{0.3824}} \\
    KADID &       & 0.3828  & \textcolor[rgb]{ 0,  0,  1}{\textbf{0.4679}} & 0.4381  & 0.4443  & \textbf{0.4640} & \textcolor[rgb]{ 1,  0,  0}{\textbf{0.5983}} &       & 0.2678  & \textbf{0.3210} & 0.3056  & 0.3099  & \textcolor[rgb]{ 0,  0,  1}{\textbf{0.3230}} & \textcolor[rgb]{ 1,  0,  0}{\textbf{0.4238}} \\
    MDLIVE &       & \textbf{0.7737} & 0.6594  & 0.5667  & \textcolor[rgb]{ 0,  0,  1}{\textbf{0.8293}} & \textcolor[rgb]{ 1,  0,  0}{\textbf{0.8647}} & 0.7579  &       & \textbf{0.5726} & 0.4738  & 0.4026  & \textcolor[rgb]{ 0,  0,  1}{\textbf{0.6391}} & \textcolor[rgb]{ 1,  0,  0}{\textbf{0.6734}} & 0.5623  \\
    MDIVL &       & 0.5361  & \textcolor[rgb]{ 0,  0,  1}{\textbf{0.6174}} & 0.5110  & 0.4750  & \textbf{0.5917} & \textcolor[rgb]{ 1,  0,  0}{\textbf{0.7890}} &       & 0.3777  & \textcolor[rgb]{ 0,  0,  1}{\textbf{0.4378}} & 0.3594  & 0.3243  & \textbf{0.4180} & \textcolor[rgb]{ 1,  0,  0}{\textbf{0.5911}} \\
    KonIQ &       & \textcolor[rgb]{ 0,  0,  1}{\textbf{0.7213}} & -     & -     & \textbf{0.6062} & -     & \textcolor[rgb]{ 1,  0,  0}{\textbf{0.7333}} &       & \textcolor[rgb]{ 0,  0,  1}{\textbf{0.5260}} & -     & -     & \textbf{0.4268} & -     & \textcolor[rgb]{ 1,  0,  0}{\textbf{0.5344}} \\
    CLIVE &       & \textbf{0.7178} & \textcolor[rgb]{ 0,  0,  1}{\textbf{0.7612}} & \textcolor[rgb]{ 1,  0,  0}{\textbf{0.8399}} & 0.5568  & 0.7216  & 0.4821  &       & 0.5293  & \textcolor[rgb]{ 0,  0,  1}{\textbf{0.5612}} & \textcolor[rgb]{ 1,  0,  0}{\textbf{0.6482}} & 0.3872  & \textbf{0.5302} & 0.3274  \\
    CID2013 &       & \textbf{0.8243} & 0.7219  & \textcolor[rgb]{ 0,  0,  1}{\textbf{0.8457}} & 0.5640  & 0.7685  & \textcolor[rgb]{ 1,  0,  0}{\textbf{0.8571}} &       & \textbf{0.6343} & 0.5385  & \textcolor[rgb]{ 0,  0,  1}{\textbf{0.6555}} & 0.3988  & 0.5784  & \textcolor[rgb]{ 1,  0,  0}{\textbf{0.6706}} \\
    SPAQ  &       & 0.6128  & \textcolor[rgb]{ 0,  0,  1}{\textbf{0.8215}} & 0.0699  & \textbf{0.7548} & \textcolor[rgb]{ 1,  0,  0}{\textbf{0.8323}} & 0.7408  &       & 0.4194  & \textcolor[rgb]{ 0,  0,  1}{\textbf{0.6112}} & 0.0439  & \textbf{0.5433} & \textcolor[rgb]{ 1,  0,  0}{\textbf{0.6227}} & 0.5347  \\
    \cmidrule{1-1}\cmidrule{3-8}\cmidrule{10-15}    $AVG_D$ &       & 0.6014  & \textbf{0.6411} & 0.5738  & 0.6025  & \textcolor[rgb]{ 0,  0,  1}{\textbf{0.6709}} & \textcolor[rgb]{ 1,  0,  0}{\textbf{0.7208}} &       & 0.4367  & \textbf{0.4618} & 0.4194  & 0.4332  & \textcolor[rgb]{ 0,  0,  1}{\textbf{0.4925}} & \textcolor[rgb]{ 1,  0,  0}{\textbf{0.5380}} \\
    $AVG_W$ &       & \textbf{0.5669} & 0.4666  & 0.2516  & \textcolor[rgb]{ 0,  0,  1}{\textbf{0.5868}} & 0.4766  & \textcolor[rgb]{ 1,  0,  0}{\textbf{0.6854}} &       & \textbf{0.4022} & 0.3365  & 0.1779  & \textcolor[rgb]{ 0,  0,  1}{\textbf{0.4171}} & 0.3474  & \textcolor[rgb]{ 1,  0,  0}{\textbf{0.4966}} \\
    \midrule
    LIVE  & \multirow{12}[6]{*}{\begin{sideways}PLCC\end{sideways}} & 0.4588  & \textbf{0.7375} & \textcolor[rgb]{ 0,  0,  1}{\textbf{0.7623}} & -     & 0.6722  & \textcolor[rgb]{ 1,  0,  0}{\textbf{0.8558}} & \multirow{12}[6]{*}{\begin{sideways}RMSE\end{sideways}} & 24.2770  & \textbf{18.4522} & \textcolor[rgb]{ 0,  0,  1}{\textbf{17.6835}} & -     & 20.2281  & \textcolor[rgb]{ 1,  0,  0}{\textbf{14.1344}} \\
    CSIQ  &       & 0.6360  & 0.5609  & \textbf{0.6549} & \textcolor[rgb]{ 0,  0,  1}{\textbf{0.7514}} & 0.6276  & \textcolor[rgb]{ 1,  0,  0}{\textbf{0.7907}} &       & 0.2026  & 0.2174  & \textbf{0.1999} & \textcolor[rgb]{ 0,  0,  1}{\textbf{0.1732}} & 0.2044  & \textcolor[rgb]{ 1,  0,  0}{\textbf{0.1607}} \\
    TID2013 &       & \textbf{0.5776} & 0.4427  & 0.4873  & \textcolor[rgb]{ 0,  0,  1}{\textbf{0.6215}} & 0.5771  & \textcolor[rgb]{ 1,  0,  0}{\textbf{0.6242}} &       & \textbf{1.0120} & 1.1116  & 1.0830  & \textcolor[rgb]{ 0,  0,  1}{\textbf{0.9711}} & 1.0124  & \textcolor[rgb]{ 1,  0,  0}{\textbf{0.9685}} \\
    KADID &       & 0.4356  & \textbf{0.4919} & 0.4788  & 0.4819  & \textcolor[rgb]{ 0,  0,  1}{\textbf{0.5035}} & \textcolor[rgb]{ 1,  0,  0}{\textbf{0.5939}} &       & 0.9745  & \textbf{0.9426} & 0.9513  & 0.9486  & \textcolor[rgb]{ 0,  0,  1}{\textbf{0.9354}} & \textcolor[rgb]{ 1,  0,  0}{\textbf{0.8710}} \\
    MDLIVE &       & 0.8160  & 0.7554  & 0.6609  & \textcolor[rgb]{ 0,  0,  1}{\textbf{0.8582}} & \textcolor[rgb]{ 1,  0,  0}{\textbf{0.8824}} & \textbf{0.8226} &       & 10.9323  & 12.3933  & 14.1928  & \textcolor[rgb]{ 0,  0,  1}{\textbf{9.7094}} & \textcolor[rgb]{ 1,  0,  0}{\textbf{8.8976}} & \textbf{10.7534} \\
    MDIVL &       & 0.5223  & \textcolor[rgb]{ 0,  0,  1}{\textbf{0.6246}} & 0.5342  & 0.4881  & \textbf{0.5798} & \textcolor[rgb]{ 1,  0,  0}{\textbf{0.7953}} &       & 20.3651  & \textcolor[rgb]{ 0,  0,  1}{\textbf{18.6496}} & 20.1880  & 20.8439  & \textbf{19.4578} & \textcolor[rgb]{ 1,  0,  0}{\textbf{14.4779}} \\
    KonIQ &       & \textcolor[rgb]{ 1,  0,  0}{\textbf{0.7259}} & -     & -     & \textbf{0.6231} & -     & \textcolor[rgb]{ 0,  0,  1}{\textbf{0.7123}} &       & \textcolor[rgb]{ 1,  0,  0}{\textbf{0.3798}} & -     & -     & \textbf{0.4319} & -     & \textcolor[rgb]{ 0,  0,  1}{\textbf{0.3876}} \\
    CLIVE &       & \textbf{0.7706} & \textcolor[rgb]{ 0,  0,  1}{\textbf{0.7739}} & \textcolor[rgb]{ 1,  0,  0}{\textbf{0.8481}} & 0.5655  & 0.7498  & 0.5364  &       & \textbf{12.9364} & \textcolor[rgb]{ 0,  0,  1}{\textbf{12.8538}} & \textcolor[rgb]{ 1,  0,  0}{\textbf{10.7528}} & 16.7398  & 13.4291  & 17.1298  \\
    CID2013 &       & \textbf{0.8261} & 0.7856  & \textcolor[rgb]{ 0,  0,  1}{\textbf{0.8326}} & 0.6077  & 0.8224  & \textcolor[rgb]{ 1,  0,  0}{\textbf{0.8717}} &       & \textbf{12.7582} & 14.0090  & \textcolor[rgb]{ 0,  0,  1}{\textbf{12.5406}} & 17.9808  & 12.8806  & \textcolor[rgb]{ 1,  0,  0}{\textbf{11.0931}} \\
    SPAQ  &       & 0.5663  & \textcolor[rgb]{ 1,  0,  0}{\textbf{0.8301}} & 0.0666  & \textbf{0.7656} & \textcolor[rgb]{ 0,  0,  1}{\textbf{0.8272}} & 0.7177  &       & 17.2262  & \textcolor[rgb]{ 1,  0,  0}{\textbf{11.6554}} & 20.8550  & \textbf{13.4464} & \textcolor[rgb]{ 0,  0,  1}{\textbf{11.7459}} & 14.5551  \\
    \cmidrule{1-1}\cmidrule{3-8}\cmidrule{10-15}    $AVG_D$ &       & 0.6335  & \textbf{0.6670} & 0.5917  & 0.6403  & \textcolor[rgb]{ 0,  0,  1}{\textbf{0.6936}} & \textcolor[rgb]{ 1,  0,  0}{\textbf{0.7321}} &       & 10.1064  & 10.0317  & 10.9385  & \textcolor[rgb]{ 0,  0,  1}{\textbf{9.0272}} & \textbf{9.8657} & \textcolor[rgb]{ 1,  0,  0}{\textbf{8.4532}} \\
    $AVG_W$ &       & \textbf{0.5852} & 0.4815  & 0.2652  & \textcolor[rgb]{ 0,  0,  1}{\textbf{0.6155}} & 0.4946  & \textcolor[rgb]{ 1,  0,  0}{\textbf{0.6803}} &       & 6.9253  & \textcolor[rgb]{ 1,  0,  0}{\textbf{5.1090}} & 7.7004  & \textbf{5.5300} & \textcolor[rgb]{ 0,  0,  1}{\textbf{5.1392}} & 5.9160  \\
    \bottomrule
    \bottomrule
    \end{tabular}%
  \label{tab:oabiqa}%
\end{table*}%

\begin{figure*}[ht]
\centering
    \subfloat[]{
			\includegraphics[width=0.31\textwidth]{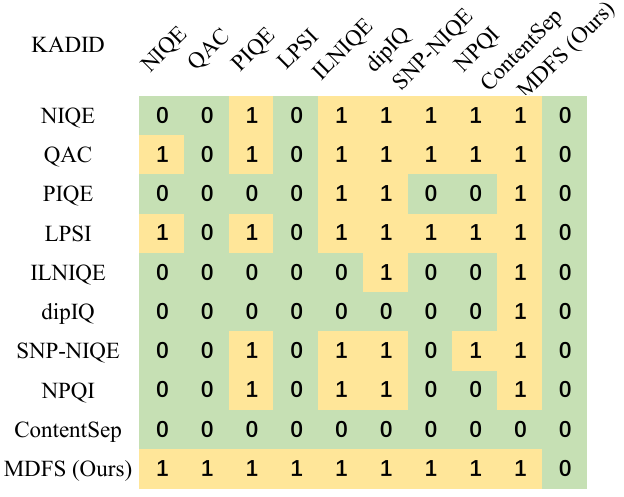}
	}
	\subfloat[]{
			\includegraphics[width=0.31\textwidth]{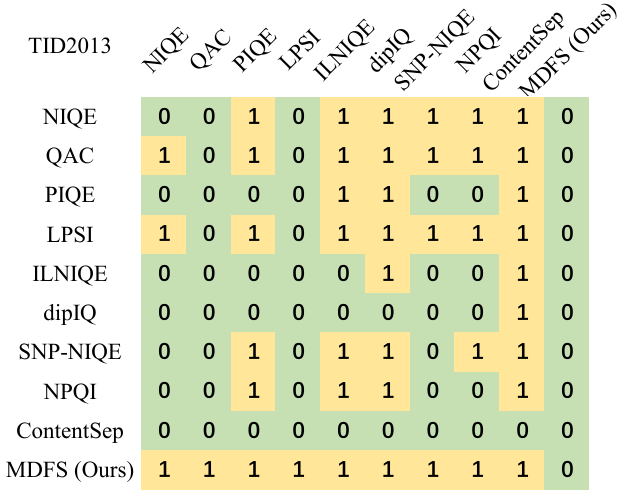}
	}
	\subfloat[]{
			\includegraphics[width=0.31\textwidth]{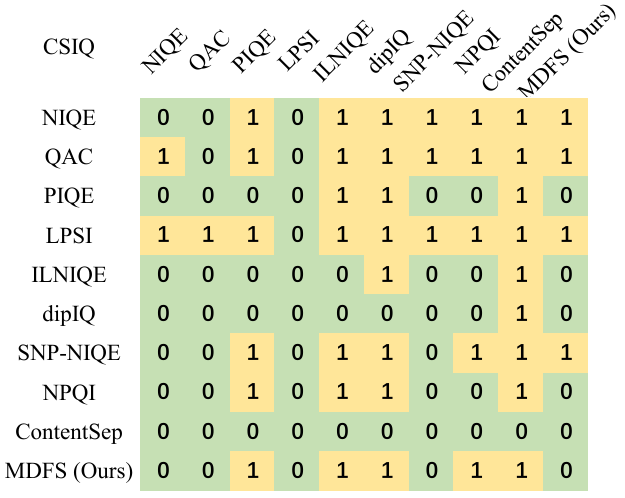}
	}
	\caption{\revise{The statistically significant test results of various OU-BIQA methods on the (a) KADID, (b) TID2013, and (c) CSIQ datasets. A value of ``1" indicates that the model in the row is significantly better than the model in the column.}
   }
	\label{fig:f_test}
\end{figure*}

\subsection{Generalizability of MDFS}
In this subsection, we evaluate the generalizability of the proposed MDFS model on four target-specific IQA datasets: underwater image (UWI), artificial intelligence generated content (AIGC), birds images (Bird), and human face images (Face). 
We compare MDFS with two well-established models, NIQE and ILNIQE, since they have relatively good performance and can be easily retrained with new high-quality images.
The experiment includes two UWI datasets: SAUD~\cite{9749233} and UWIQA~\cite{yang2021reference}, two AIGC datasets: AGIQA~\cite{li2023agiqa} and AIGCIQA~\cite{wang2023aigciqa2023}, one Bird dataset: MMQA-Birds~\cite{yu2021blind}, and one Face dataset: GFIQA-20k~\cite{gfiqa20k} for testing. 
Specifically, for high-quality image dataset for retraining, we select the first 200 high-quality enhanced images from UID2021~\cite{hou2023uid2021} as the high-quality UWI images, 200 high-quality bird images from the Caltech-UCSD Birds-200-2011 dataset~\cite{wah2011caltech}, and 200 high-quality face images from the CelebA-HQ~\cite{huang2018introvae} dataset.
Notably, the high-quality image dataset for retraining has a distribution similar to the corresponding testing dataset, but there is no overlap between the images used for training and testing. This ensures that the performance of the model is evaluated on unseen data, testing its ability to generalize to new and unobserved images.

From the results presented in Table~\ref{tab:generalizability}, one can observe that the retrained model generally outperforms the original one in most cases.
This improvement can be attributed to the fact that the retrained model is learned from target-specific images that are close to the testing images, whereas the original model is trained on general high-quality images. \revise{In comparison to the noticeable improvement in performance after retraining on two tasks (i.e., bird and underwater datasets), the results of the model retrained on the face dataset remain remarkably close to the original results. We speculate that this close resemblance may be attributed to the saturation of model performance on this task.} Additionally, the retrained MDFS model demonstrates superior performance compared to other methods across all three image categories, showcasing the effectiveness of the proposed MDFS framework in generalizing to diverse target-specific IQA tasks.

\subsection{Significant Test}
\label{sec:significant}
In this subsection, the F-test is adopted as statistical analysis to illustrate the superiority of the proposed MDFS model compared with other OU-BIQA models.
The results of the F-test between each pair of BIQA models on KADID, TID2013, and CSIQ datasets are presented in Figure~\ref{fig:f_test}. 
A value of ``1" (highlighted in yellow) indicates that the model in the row significantly outperforms the model in the column, while a value of ``0" (highlighted in green) indicates that both algorithms in the column and row are statistically equivalent. 
The results depicted in Figure~\ref{fig:f_test} reveal that the proposed MDFS outperforms all the classical and state-of-the-art OU-BIQA models on both the KADID and TID2013 datasets. 
On the CSIQ dataset, MDFS demonstrates superiority over most of the comparison OU-IQA methods (\textit{i.e.}, PIQE, ILNIQE, dipIQ, NPQI, and ContentSep). 
These findings highlight the statistical significance of the performance advantage of MDFS on different datasets.

\subsection{Cross-dataset Comparison with OA-BIQA Methods}
In this subsection, we conduct a comprehensive comparison of the proposed MDFS with five OA-BIQA methods, including PaQ2PiQ~\cite{ying2020patches}, HyperIQA~\cite{su2020blindly}, MANIQA~\cite{yang2022maniqa}, VCRNet~\cite{pan2022vcrnet}, and MUSIQ~\cite{ke2021musiq}.
From the results in Table~\ref{tab:oabiqa}, one can observe that MDFS outperforms the compared OA-BIQA methods on ten public datasets in most cases. 
Furthermore, both the direct average score and the weighted average score of the proposed MDFS demonstrate superior performance compared to all OA-BIQA methods in terms of SROCC, KROCC, and PLCC. 
This indicates that MDFS exhibits remarkable performance compared to algorithms that require training based on subjective scores. 
These findings underscore the excellence of the proposed MDFS approach and its suitability for a wide range of IQA applications. 
\revise{We believe this is mainly due to the fact that our algorithm takes advantage of the visual model pre-trained on a large data set, which enables it to extract more general features and conduct statistical analysis.}

\begin{table}[t]
  \centering
  \caption{Computational efficiency compared with state-of-the-art methods on the CID2013 and TID2013 datasets. 
  }
   \begin{tabular}{cccc}
    \toprule
    \toprule
    Methods & \multicolumn{1}{p{10.0 em}}{Programming Language} & CID2013 & TID2013 \\
    \midrule
    NIQE  & MATLAB & 0.2216  & \textcolor[rgb]{ 0,  0,  1}{\textbf{0.0263}} \\
    QAC   & MATLAB & 0.4259  & 0.0618  \\
    PIQE  & MATLAB & 0.3502  & 0.0413  \\
    LPSI  & MATLAB & \textbf{0.1304} & \textcolor[rgb]{ 1,  0,  0}{\textbf{0.0194}} \\
    ILNIQE & MATLAB & 1.8163  & 1.5908  \\
    dipIQ & MATLAB & 1.8800  & 0.9905  \\
    SNP-NIQE & MATLAB & 11.7991  & 0.8245  \\
    NPQI  & MATLAB & 7.8984  & 0.8534  \\
    ContentSep & Python & \textcolor[rgb]{ 1,  0,  0}{\textbf{0.0470}} & 0.1518  \\
    MDFS (Ours) & Python & \textcolor[rgb]{ 0,  0,  1}{\textbf{0.0558}} & \textbf{0.0289} \\
    \bottomrule
    \bottomrule
    \end{tabular}%
  \label{tab:computational_efficiency}%
\end{table}%

\subsection{\revise{Computational Efficiency}}
\label{sec:efficiency}
\revise{In this subsection, we conduct the computational efficiency comparison experiment, which is an important factor in evaluating the performance of IQA models in real-world applications. The computational efficiency of all BIQA models is assessed using the CID2013 (474 distorted images with a resolution of $1600\times1200$) and TID2013 (3000 distorted images with a resolution of $512\times384$) datasets, which have varying image resolutions, to compute the average running time per image. 
All the experiments run on the computer with an Intel i7-9700K CPU @ 3.60GHz and an NVIDIA GeForce RTX 2080 Ti GPU. 
Furthermore, all the codes of IQA models are performed under the same suggestion of the corresponding authors. The results in Table~\ref{tab:computational_efficiency} indicate that the proposed MDFS model demonstrates similar computational speed to the fastest algorithms across the two datasets. 
Moreover, the proposed model manifests pronounced advantages specifically in the context of high-resolution imagery.}

\begin{table*}[t]
  \centering
  \setlength{\tabcolsep}{5pt}
  \caption{Ablation studies on various training datasets. The top three are marked in bold~\textcolor[rgb]{ 1,  0,  0}{\textbf{red}}, \textcolor[rgb]{ 0,  0,  1}{\textbf{blue}}, and~\textbf{black}, respectively.}
    \begin{tabular}{cc|cccccccccc}
    \toprule
    \toprule
    \multicolumn{1}{p{6.815em}}{Training dataset} & Criteria & LIVE  & CSIQ  & TID2013 & KADID & MDLIVE & MDIVL & KonIQ & CLIVE & CID2013 & SPAQ \\
    \midrule
    \multirow{4}[2]{*}{Waterloo} & SROCC & \textbf{0.9072} & \textcolor[rgb]{ 0,  0,  1}{\textbf{0.7964}} & \textcolor[rgb]{ 1,  0,  0}{\textbf{0.5752}} & \textcolor[rgb]{ 0,  0,  1}{\textbf{0.6214}} & \textcolor[rgb]{ 1,  0,  0}{\textbf{0.7763}} & \textcolor[rgb]{ 1,  0,  0}{\textbf{0.8331}} & 0.4542  & 0.3998  & \textcolor[rgb]{ 0,  0,  1}{\textbf{0.6249}} & \textbf{0.7191} \\
          & KROCC & \textbf{0.7296} & \textcolor[rgb]{ 0,  0,  1}{\textbf{0.5998}} & \textcolor[rgb]{ 1,  0,  0}{\textbf{0.4130}} & \textcolor[rgb]{ 0,  0,  1}{\textbf{0.4431}} & \textcolor[rgb]{ 1,  0,  0}{\textbf{0.5644}} & \textcolor[rgb]{ 1,  0,  0}{\textbf{0.6332}} & 0.3114  & 0.2705  & \textcolor[rgb]{ 0,  0,  1}{\textbf{0.4557}} & \textbf{0.5195} \\
          & PLCC  & 0.7925  & \textcolor[rgb]{ 0,  0,  1}{\textbf{0.7980}} & \textcolor[rgb]{ 1,  0,  0}{\textbf{0.6380}} & \textcolor[rgb]{ 0,  0,  1}{\textbf{0.6434}} & \textcolor[rgb]{ 0,  0,  1}{\textbf{0.7212}} & \textcolor[rgb]{ 0,  0,  1}{\textbf{0.7392}} & 0.4687  & 0.4659  & \textbf{0.5631} & \textbf{0.6605} \\
          & RMSE  & 16.6633  & \textcolor[rgb]{ 0,  0,  1}{\textbf{0.1582}} & \textcolor[rgb]{ 1,  0,  0}{\textbf{0.9546}} & \textcolor[rgb]{ 0,  0,  1}{\textbf{0.8288}} & \textcolor[rgb]{ 0,  0,  1}{\textbf{13.1012}} & \textcolor[rgb]{ 0,  0,  1}{\textbf{16.0835}} & 0.4878  & 17.9591  & \textbf{18.7101} & \textbf{15.6932} \\
    \midrule
    \multirow{4}[2]{*}{D-NIQE} & SROCC & 0.8900  & \textbf{0.7792} & \textcolor[rgb]{ 0,  0,  1}{\textbf{0.5524}} & \textcolor[rgb]{ 1,  0,  0}{\textbf{0.6299}} & 0.7245  & 0.6629  & \textbf{0.4721} & \textbf{0.4404} & 0.4894  & 0.6897  \\
          & KROCC & 0.7052  & \textbf{0.5856} & \textcolor[rgb]{ 0,  0,  1}{\textbf{0.3965}} & \textcolor[rgb]{ 1,  0,  0}{\textbf{0.4509}} & 0.5183  & 0.4705  & \textbf{0.3248} & \textbf{0.3012} & 0.3504  & 0.4926  \\
          & PLCC  & \textbf{0.7998} & 0.7859  & \textbf{0.6266} & \textcolor[rgb]{ 1,  0,  0}{\textbf{0.6541}} & 0.6189  & 0.5388  & \textbf{0.4709} & \textbf{0.4711} & 0.4366  & 0.6418  \\
          & RMSE  & \textbf{16.3997} & 0.1623  & \textbf{0.9662} & \textcolor[rgb]{ 1,  0,  0}{\textbf{0.8189}} & 14.8549  & 20.1192  & \textbf{0.4871} & \textbf{17.9028} & 20.3681  & 16.0279  \\
    \midrule
    \multirow{4}[2]{*}{KADID} & SROCC & \textcolor[rgb]{ 0,  0,  1}{\textbf{0.9093}} & \textcolor[rgb]{ 1,  0,  0}{\textbf{0.8101}} & \textbf{0.5507} & -     & \textcolor[rgb]{ 0,  0,  1}{\textbf{0.7597}} & \textbf{0.7352} & \textcolor[rgb]{ 0,  0,  1}{\textbf{0.5535}} & \textcolor[rgb]{ 1,  0,  0}{\textbf{0.5313}} & \textbf{0.5996} & \textcolor[rgb]{ 0,  0,  1}{\textbf{0.7387}} \\
          & KROCC & \textcolor[rgb]{ 0,  0,  1}{\textbf{0.7331}} & \textcolor[rgb]{ 1,  0,  0}{\textbf{0.6165}} & \textbf{0.3937} & -     & \textbf{0.5550} & \textbf{0.5351} & \textcolor[rgb]{ 0,  0,  1}{\textbf{0.3893}} & \textcolor[rgb]{ 1,  0,  0}{\textbf{0.3666}} & \textbf{0.4428} & \textcolor[rgb]{ 1,  0,  0}{\textbf{0.5374}} \\
          & PLCC  & \textcolor[rgb]{ 1,  0,  0}{\textbf{0.9127}} & \textcolor[rgb]{ 1,  0,  0}{\textbf{0.8161}} & \textcolor[rgb]{ 0,  0,  1}{\textbf{0.6327}} & -     & \textbf{0.7126} & \textbf{0.7264} & \textcolor[rgb]{ 0,  0,  1}{\textbf{0.4949}} & \textcolor[rgb]{ 1,  0,  0}{\textbf{0.5456}} & \textcolor[rgb]{ 0,  0,  1}{\textbf{0.6432}} & \textcolor[rgb]{ 0,  0,  1}{\textbf{0.6821}} \\
          & RMSE  & \textcolor[rgb]{ 1,  0,  0}{\textbf{11.1664}} & \textcolor[rgb]{ 1,  0,  0}{\textbf{0.1519}} & \textcolor[rgb]{ 0,  0,  1}{\textbf{0.9600}} & -     & \textbf{13.2674} & \textbf{16.4126} & \textcolor[rgb]{ 0,  0,  1}{\textbf{0.4798}} & \textcolor[rgb]{ 1,  0,  0}{\textbf{17.0093}} & \textcolor[rgb]{ 0,  0,  1}{\textbf{17.3358}} & \textcolor[rgb]{ 0,  0,  1}{\textbf{15.2846}} \\
    \midrule
    \multirow{4}[2]{*}{DIV2K (Ours)} & SROCC & \textcolor[rgb]{ 1,  0,  0}{\textbf{0.9361}} & 0.7774  & 0.5363  & \textbf{0.5983} & \textbf{0.7579} & \textcolor[rgb]{ 0,  0,  1}{\textbf{0.7890}} & \textcolor[rgb]{ 1,  0,  0}{\textbf{0.7333}} & \textcolor[rgb]{ 0,  0,  1}{\textbf{0.4821}} & \textcolor[rgb]{ 1,  0,  0}{\textbf{0.8571}} & \textcolor[rgb]{ 1,  0,  0}{\textbf{0.7408}} \\
          & KROCC & \textcolor[rgb]{ 1,  0,  0}{\textbf{0.7709}} & 0.5823  & 0.3824  & \textbf{0.4238} & \textcolor[rgb]{ 0,  0,  1}{\textbf{0.5623}} & \textcolor[rgb]{ 0,  0,  1}{\textbf{0.5911}} & \textcolor[rgb]{ 1,  0,  0}{\textbf{0.5344}} & \textcolor[rgb]{ 0,  0,  1}{\textbf{0.3274}} & \textcolor[rgb]{ 1,  0,  0}{\textbf{0.6706}} & \textcolor[rgb]{ 0,  0,  1}{\textbf{0.5347}} \\
          & PLCC  & \textcolor[rgb]{ 0,  0,  1}{\textbf{0.8558}} & \textbf{0.7907} & 0.6242  & \textbf{0.5939} & \textcolor[rgb]{ 1,  0,  0}{\textbf{0.8226}} & \textcolor[rgb]{ 1,  0,  0}{\textbf{0.7953}} & \textcolor[rgb]{ 1,  0,  0}{\textbf{0.7123}} & \textcolor[rgb]{ 0,  0,  1}{\textbf{0.5364}} & \textcolor[rgb]{ 1,  0,  0}{\textbf{0.8717}} & \textcolor[rgb]{ 1,  0,  0}{\textbf{0.7177}} \\
          & RMSE  & \textcolor[rgb]{ 0,  0,  1}{\textbf{14.1344}} & \textbf{0.1607} & 0.9685  & \textbf{0.8710} & \textcolor[rgb]{ 1,  0,  0}{\textbf{10.7534}} & \textcolor[rgb]{ 1,  0,  0}{\textbf{14.4779}} & \textcolor[rgb]{ 1,  0,  0}{\textbf{0.3876}} & \textcolor[rgb]{ 0,  0,  1}{\textbf{17.1298}} & \textcolor[rgb]{ 1,  0,  0}{\textbf{11.0931}} & \textcolor[rgb]{ 1,  0,  0}{\textbf{14.5551}} \\
    \bottomrule
    \bottomrule
    \end{tabular}%
  \label{tab:high-quality}%
\end{table*}%

\begin{table}[t]
  \centering
  \setlength{\tabcolsep}{0.8pt}
  \caption{Ablation studies on various network backbones. 
 }
    \begin{tabular}{cc|cccccrc}
    \toprule
    \toprule
     Datasets & Criteria & ConvNet & Inception & PNAS  & ResNet & VGG   & \multicolumn{1}{c}{ViT} & EN \\
    \midrule
    \multirow{4}[1]{*}{{\begin{sideways}LIVE\end{sideways}} } & SROCC & 0.8138  & 0.4559  & 0.8871  & \textbf{0.8872} & \textcolor[rgb]{ 0,  0,  1}{\textbf{0.8896}} & 0.4675 & \textcolor[rgb]{ 1,  0,  0}{\textbf{0.9361}} \\
          & KROCC & 0.5991  & 0.3206  & \textcolor[rgb]{ 0,  0,  1}{\textbf{0.6995}} & 0.6887  & \textbf{0.6943} & 0.3193 & \textcolor[rgb]{ 1,  0,  0}{\textbf{0.7709}} \\
          & RMSE  & 15.7004  & 23.0483  & 16.3517  & \textcolor[rgb]{ 0,  0,  1}{\textbf{12.3457}} & \textcolor[rgb]{ 1,  0,  0}{\textbf{12.1704}} & 24.2031 & \textbf{14.1344} \\
          & PLCC  & 0.8184  & 0.5370  & 0.8011  & \textcolor[rgb]{ 0,  0,  1}{\textbf{0.8921}} & \textcolor[rgb]{ 1,  0,  0}{\textbf{0.8953}} & 0.464 & \textbf{0.8558} \\
    \midrule
    \multirow{4}[1]{*}{{\begin{sideways}TID2013\end{sideways}} } & SROCC & 0.4106  & 0.2559  & \textcolor[rgb]{ 0,  0,  1}{\textbf{0.5010}} & \textbf{0.4495} & 0.4160  & 0.3403 & \textcolor[rgb]{ 1,  0,  0}{\textbf{0.5363}} \\
          & KROCC & 0.2756  & 0.1738  & \textcolor[rgb]{ 0,  0,  1}{\textbf{0.3491}} & \textbf{0.3085} & 0.2857  & 0.2318 & \textcolor[rgb]{ 1,  0,  0}{\textbf{0.3824}} \\
          & RMSE  & 1.1332  & 1.1526  & \textbf{1.0550} & 1.0866  & \textcolor[rgb]{ 0,  0,  1}{\textbf{1.0418}} & 1.1345 & \textcolor[rgb]{ 1,  0,  0}{\textbf{0.9685}} \\
          & PLCC  & 0.4054  & 0.3682  & \textbf{0.5251} & 0.4813  & \textcolor[rgb]{ 0,  0,  1}{\textbf{0.5420}} & 0.4031 & \textcolor[rgb]{ 1,  0,  0}{\textbf{0.6242}} \\
    \bottomrule
    \bottomrule
    \end{tabular}
  \label{tab:ab_network}
\end{table}

\subsection{Ablation Studies}
\label{sec:ablation}
This subsection conducts extensive ablation studies to evaluate the impact of each component in the proposed MDFS model, including training dataset, network backbones, window size, quality calculation, and contrast feature.

\subsubsection{Training Dataset}
In our study, we utilize the DIV2K dataset for training purposes due to its excellent image quality and diverse content. 
However, we also explored the performance of using other high-quality image datasets, specifically the Waterloo dataset, the reference images of KADID, and the training dataset of NIQE (referred to as D-NIQE). 
The evaluation results are summarized in Table~\ref{tab:high-quality}. 
It is evident that training on the DIV2K dataset led to relatively better performance compared to training on the other datasets. 
Furthermore, models trained on those alternative datasets still delivered commendable results when compared to existing algorithms, as demonstrated in Table~\ref{tab:overall}. This highlights the robustness and versatility of the proposed MDFS model.

\subsubsection{Network Backbone}
In our study, we explored the use of various existing networks as the backbone for the MDFS to provide feature maps, including  VGG~\cite{simonyan2014very}, ResNet~\cite{he2016deep}, PNAS~\cite{liu2018progressive}, ConvNet~\cite{liu2022convnet}, ViT~\cite{dosovitskiy2020image}, and EfficientNet~\cite{tan2019efficientnet}. 
Table~\ref{tab:ab_network} presents the SROCC results on the LIVE and TID2013 datasets with various datasets.
These results indicate that utilizing alternative backbone networks, such as ResNet, VGG, PNAS, and ViT did not yield results as promising as when employing EfficientNet.
Therefore, we use EfficientNet as the preferred backbone for the feature extraction module as it consistently shows the most promising results.

\begin{table}[t]
  \centering
  \caption{Ablation study on various window sizes. 
  }
    \begin{tabular}{cc|cccc}
    \toprule
    \toprule
    Datasets & Criteria & $s_w$=3 & $s_w$=5 & $s_w$=7 & Ours \\
    \midrule
    \multirow{4}[2]{*}{CID2013} & SROCC & 0.8410  & \textbf{0.8489} & \textcolor[rgb]{ 0,  0,  1}{\textbf{0.8495}} & \textcolor[rgb]{ 1,  0,  0}{\textbf{0.8571}} \\
          & KROCC & 0.6517  & \textbf{0.6606} & \textcolor[rgb]{ 0,  0,  1}{\textbf{0.6617}} & \textcolor[rgb]{ 1,  0,  0}{\textbf{0.6706}} \\
          & RMSE  & \textcolor[rgb]{ 0,  0,  1}{\textbf{11.6844}} & \textbf{13.7513} & 13.7975  & \textcolor[rgb]{ 1,  0,  0}{\textbf{11.0931}} \\
          & PLCC  & \textcolor[rgb]{ 0,  0,  1}{\textbf{0.8565}} & \textbf{0.7944} & 0.7928  & \textcolor[rgb]{ 1,  0,  0}{\textbf{0.8717}} \\
    \midrule
    \multirow{4}[2]{*}{LIVE} & SROCC & 0.9337  & \textcolor[rgb]{ 0,  0,  1}{\textbf{0.9352}} & \textbf{0.9346} & \textcolor[rgb]{ 1,  0,  0}{\textbf{0.9361}} \\
          & KROCC & 0.7665  & \textcolor[rgb]{ 0,  0,  1}{\textbf{0.7685}} & \textbf{0.7669} & \textcolor[rgb]{ 1,  0,  0}{\textbf{0.7709}} \\
          & RMSE  & \textcolor[rgb]{ 0,  0,  1}{\textbf{9.9014}} & \textbf{12.8541} & \textcolor[rgb]{ 1,  0,  0}{\textbf{9.8584}} & 14.1344  \\
          & PLCC  & \textcolor[rgb]{ 0,  0,  1}{\textbf{0.9320}} & \textbf{0.8824} & \textcolor[rgb]{ 1,  0,  0}{\textbf{0.9326}} & 0.8558  \\
    \bottomrule
    \bottomrule
    \end{tabular}%
  \label{tab:ab_window_size}%
\end{table}%

\subsubsection{Window Size}
In our proposed MDFS model, we introduce a novel window size calculation method, as described in Equ. (\ref{eq:win}), which dynamically adjusts the window size based on the dimensions of the input image. 
To evaluate its effectiveness, we compare the proposed dynamic window size with three fixed window sizes: 3, 5, and 7.
\revise{The results in Table~\ref{tab:ab_window_size} demonstrate that the proposed dynamic window algorithm not only enhances the accuracy of the proposed MDFS model but also exhibits robustness across various input image sizes}. 
This indicates that our approach adapts well to images of different dimensions, making it a versatile and effective solution.

\subsubsection{\revise{Contrast Feature}}
\revise{In our proposed MDFS model, we use the standard deviation as a measure of contrast to extract HVS-sensitive information. Herein, we investigate the performance of various contrast information extraction methods. To be specific, \textit{w/o. std} represents the MDFS model without weighting map (\textit{i.e.}, standard deviation); \textit{w. var} refers to the MDFS with variance as weighting map; \textit{w. entropy} indicates the MDFS with entropy feature as weighting map; and \textit{w. $\bm{w'}$} denotes using a new weighting map $\bm{w'}$ defined as follows:}
\revise{\begin{equation}
    \bm{w'} = 1 / ({1+e^{-\bm{F}_\sigma^{2\sigma}/(\bm{F}_\sigma^\mu + \delta)}}),
\end{equation}
where $\bm{F}_\sigma^\mu $ and $\bm{F}_\sigma^\sigma $ represent the mean and standard deviation of the $\bm{F}_\sigma$, respectively. $\delta $ is a small positive number (\textit{e.g.}, $\delta=1 \times e^{-12}$) to prevent the denominator from being zero. The results in Table~\ref{tab:ablation_contrast} demonstrate that using standard deviation in the MDFS yields superior performance compared to other variants. The potential reason may be that the standard deviation can better capture HVS-sensitive features.}

\begin{table}[t]
  \centering
  \caption{Ablation study on different contrast features. 
  }
  \setlength{\tabcolsep}{4pt}
    \begin{tabular}{cc|ccccc}
    \toprule
    \toprule
          & \multicolumn{1}{c|}{Datasets} & w/o. std  & w. var   & w. entropy   & w. $w'$ & w. std (ours) \\
    \midrule 
    \multirow{9}[5]{*}{\begin{sideways}SROCC\end{sideways}} 
   & LIVE  & 0.9121 & \textbf{0.9136} & 0.9124 & \textcolor[rgb]{ 0,  0,  1}{\textbf{0.9136}} & \textcolor[rgb]{ 1,  0,  0}{\textbf{0.9361}} \\
   & CSIQ  & 0.7321 & \textcolor[rgb]{ 0,  0,  1}{\textbf{0.7333}} & 0.7308 & \textbf{0.7333} & \textcolor[rgb]{ 1,  0,  0}{\textbf{0.7774}} \\
   & TID2013 & 0.5181 & \textcolor[rgb]{ 0,  0,  1}{\textbf{0.5226}} & 0.5213 & \textbf{0.5226} & \textcolor[rgb]{ 1,  0,  0}{\textbf{0.5363}} \\
   & KADID & 0.5716 & \textcolor[rgb]{ 0,  0,  1}{\textbf{0.5828}} & 0.5713 & \textbf{0.5820} & \textcolor[rgb]{ 1,  0,  0}{\textbf{0.5983}} \\
   & MDLIVE & 0.6702 & \textcolor[rgb]{ 0,  0,  1}{\textbf{0.6902}} & 0.6761 & \textbf{0.6871} & \textcolor[rgb]{ 1,  0,  0}{\textbf{0.7579}} \\
   & MDIVL & \textcolor[rgb]{ 1,  0,  0}{\textbf{0.8008}} & 0.7971 & \textcolor[rgb]{ 0,  0,  1}{\textbf{0.7993}} & \textbf{0.7976} & 0.7890 \\
   & KonIQ & 0.5778 & \textcolor[rgb]{ 0,  0,  1}{\textbf{0.5842}} & 0.5796 & \textbf{0.5838} & \textcolor[rgb]{ 1,  0,  0}{\textbf{0.7333}} \\
   & CLIVE & 0.3596 & \textbf{0.3655} & \textcolor[rgb]{ 0,  0,  1}{\textbf{0.3682}} & 0.3648 & \textcolor[rgb]{ 1,  0,  0}{\textbf{0.4821}} \\
   & CID2013 & \textcolor[rgb]{ 0,  0,  1}{\textbf{0.5405}} & 0.5370 & 0.5288 & \textbf{0.5395} & \textcolor[rgb]{ 1,  0,  0}{\textbf{0.8571}} \\
   & SPAQ  & 0.5626 & \textcolor[rgb]{ 0,  0,  1}{\textbf{0.6007}} & 0.5765 & \textbf{0.5974} & \textcolor[rgb]{ 1,  0,  0}{\textbf{0.7408}} \\
    \bottomrule
    \bottomrule
    \end{tabular}%
  \label{tab:ablation_contrast}%
\end{table}%

\begin{table}[t]
  \centering
  \caption{Ablation study on different distance calculation methods. 
  }
  \setlength{\tabcolsep}{4pt}
    \begin{tabular}{cc|ccccc}
    \toprule
    \toprule
          & \multicolumn{1}{c|}{Datasets} & MMD  & EMD   & SWD   & KL & MDFS (Ours) \\
    \midrule 
    \multirow{9}[5]{*}{\begin{sideways}SROCC\end{sideways}} 
   & LIVE  & 0.4710  & \textcolor[rgb]{ 0,  0,  1}{\textbf{0.6531}} & 0.3808  & \textbf{0.6497} & \textcolor[rgb]{ 1,  0,  0}{\textbf{0.9361}} \\
   &  CSIQ  & 0.2427  & \textcolor[rgb]{ 0,  0,  1}{\textbf{0.4740}} & 0.1930  & \textbf{0.4662} & \textcolor[rgb]{ 1,  0,  0}{\textbf{0.7774}} \\
   &  TID2013 & \textcolor[rgb]{ 0,  0,  1}{\textbf{0.2583}} & 0.2379  & 0.0693  & \textbf{0.2340} & \textcolor[rgb]{ 1,  0,  0}{\textbf{0.5363}} \\
   &  KADID & 0.2202  & \textcolor[rgb]{ 0,  0,  1}{\textbf{0.2602}} & 0.1368  & \textbf{0.2520} & \textcolor[rgb]{ 1,  0,  0}{\textbf{0.5983}} \\
   &  MDLIVE & \textcolor[rgb]{ 0,  0,  1}{\textbf{0.3429}} & 0.3301  & 0.2052  & \textbf{0.3343} & \textcolor[rgb]{ 1,  0,  0}{\textbf{0.7579}} \\
   &  MDIVL & 0.3455  & \textbf{0.4510} & 0.0007  & \textcolor[rgb]{ 0,  0,  1}{\textbf{0.4516}} & \textcolor[rgb]{ 1,  0,  0}{\textbf{0.7890}} \\
   &  KonIQ & 0.2458  & \textcolor[rgb]{ 0,  0,  1}{\textbf{0.3239}} & 0.0963  & \textbf{0.3182} & \textcolor[rgb]{ 1,  0,  0}{\textbf{0.7333}} \\
   &  CLIVE & 0.0501  & \textcolor[rgb]{ 0,  0,  1}{\textbf{0.1873}} & 0.0205  & \textbf{0.1791} & \textcolor[rgb]{ 1,  0,  0}{\textbf{0.4821}} \\
   &  CID2013 & 0.0187  & \textbf{0.1980} & \textcolor[rgb]{ 0,  0,  1}{\textbf{0.3871}} & 0.1952  & \textcolor[rgb]{ 1,  0,  0}{\textbf{0.8571}} \\
   &  SPAQ  & 0.3828  & \textcolor[rgb]{ 0,  0,  1}{\textbf{0.5144}} & 0.3739  & \textbf{0.5105} & \textcolor[rgb]{ 1,  0,  0}{\textbf{0.7408}} \\
    \bottomrule
    \bottomrule
    \end{tabular}%
  \label{tab:ablation_distance}%
\end{table}%

\subsubsection{\revise{Quality Calculation}}
\revise{We explored the performance of various distance algorithms to calculate the final quality score, including Maximum Mean Discrepancy (MMD)~\cite{gretton2006kernel}, Earth Mover’s distance (EMD)~\cite{rubner1998metric}, Sliced Wasserstein Distance (SWD)~\cite{kolouri2019generalized}, and Kullback-Leibler (KL) Divergence~\cite{kullback1951information}. From the results in Table~\ref{tab:ablation_distance}, it is evident that employing the MVG distance yields superior and robust results compared with other distance algorithms.}

\begin{figure}[t]
\centering
    \subfloat[]{\includegraphics[width=0.22\textwidth]{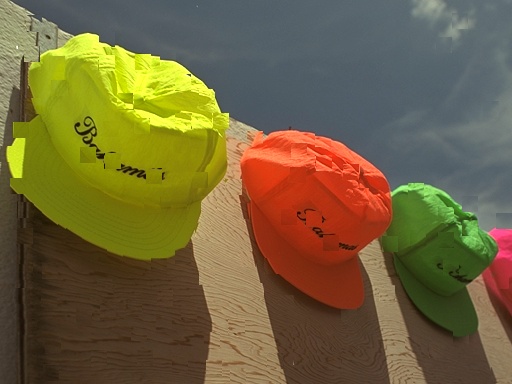}}\hfil%
    \subfloat[]{\includegraphics[width=0.22\textwidth]{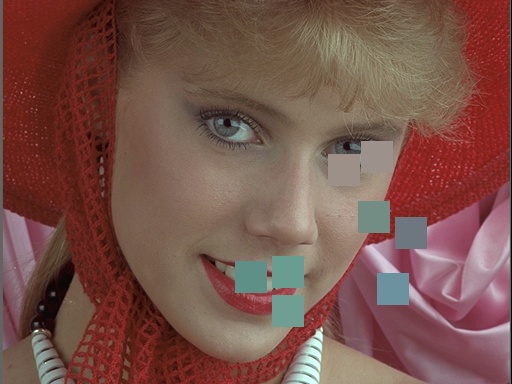}} 
\caption{\revise{Failure cases. The distorted images are generated by the (a) NEPN and (b) LBD distortions in TID2013.}}
\label{fig:limitation}
\end{figure}

\subsection{\revise{Limitation}}
\revise{The proposed MDFS model undoubtedly presents significant advancements in the OU-BIQA task. However, it also comes with certain limitations. Firstly, the proposed method relies on the high-quality image dataset for the learning of reference features. To efficiently identify the difference between high-quality images and distorted images, both the high visual quality and content diversity are required for the images used for learning. Secondly, the proposed method calculates the quality score based on the statistical feature analysis of the overall image, therefore ignoring the rationality of the local patches. For instance, NEPN and LBD distortions alter local patch content but have minimal impact on overall statistical features, as illustrated in Figure~\ref{fig:limitation}. Consequently, MDFS demonstrates poor performance on images with these distortion types, as evidenced in Table~\ref{tab:distortion_type}. In future work, it may be worthwhile to enhance the accuracy of the BIQA model on these distortion types by incorporating the local positional features of the images.}

\section{Conclusion}
\label{sec:conclusion}
\revise{In this paper, we introduce a novel opinion-unaware blind image quality assessment (OU-BIQA) model called the Multi-scale Deep Feature Statistic (MDFS) model, which eliminates the need for human-rated data during training. The core idea of our approach involves integrating multi-scale deep features with a traditional statistical analysis model for OU-BIQA. 
On one hand, deep features provide richer and more expressive representations compared to conventional features. 
On the other hand, the statistical analysis model is highly efficient and stable, making the training process more cost-effective. 
Experimental results across various datasets demonstrate that our model achieves superior consistency with human visual perception compared to existing BIQA methods, while also exhibiting improved generalizability across diverse target-specific BIQA tasks.}

\revise{While our research has made significant strides, it is imperative to acknowledge the limitation of some distortion types that do not significantly impact the global statistical data of the image. This results in a subpar performance of most IQA metrics on these distortion types. Future research endeavors could explore novel methodologies to address this limitation, potentially involving finer-grained distortion analysis based on local information or tailored processing strategies for different distortion types.}

\bibliography{reference} 
\bibliographystyle{IEEEtran}

\end{document}